\documentclass[english]{article}
\usepackage[T1]{fontenc}
\usepackage[latin9]{inputenc}
\usepackage{array}
\usepackage{float}
\usepackage{textcomp}
\usepackage{url}
\usepackage{multirow}
\usepackage{amsmath}
\usepackage{amsthm}
\usepackage{amssymb}
\usepackage{graphicx}
\usepackage{wasysym}

\makeatletter

\newcommand{\lyxmathsym}[1]{\ifmmode\begingroup\def\b@ld{bold}
  \text{\ifx\math@version\b@ld\bfseries\fi#1}\endgroup\else#1\fi}

\providecommand{\tabularnewline}{\\}
\floatstyle{ruled}
\newfloat{algorithm}{tbp}{loa}
\providecommand{\algorithmname}{Algorithm}
\floatname{algorithm}{\protect\algorithmname}

\usepackage[T1]{fontenc}
\usepackage{aaai22}  % DO NOT CHANGE THIS
\usepackage{times}  % DO NOT CHANGE THIS
\usepackage{helvet}  % DO NOT CHANGE THIS
\usepackage{courier}  % DO NOT CHANGE THIS
\usepackage{graphicx} % DO NOT CHANGE THIS
\usepackage{natbib}  % DO NOT CHANGE THIS AND DO NOT ADD ANY OPTIONS TO IT
\usepackage{caption} % DO NOT CHANGE THIS AND DO NOT ADD ANY OPTIONS TO IT
\DeclareCaptionStyle{ruled}{labelfont=normalfont,labelsep=colon,strut=off} % DO NOT CHANGE THIS
\usepackage{algorithm}
\usepackage{algorithmic}

\makeatother

\usepackage{babel}
\begin{document}
\title{Episodic Policy Gradient Training}
\author{Hung Le, Majid Abdolshah, Thommen K. George, Kien Do, Dung Nguyen,
Svetha Venkatesh{\small{}}\\
\normalfont{Applied AI Institute, Deakin University, Geelong, Australia}\\
\normalfont\texttt{thai.le@deakin.edu.au}}
\maketitle
\begin{abstract}
We introduce a novel training procedure for policy gradient methods
wherein episodic memory is used to optimize the hyperparameters of
reinforcement learning algorithms on-the-fly. Unlike other hyperparameter
searches, we formulate hyperparameter scheduling as a standard Markov
Decision Process and use episodic memory to store the outcome of used
hyperparameters and their training contexts. At any policy update
step, the policy learner refers to the stored experiences, and adaptively
reconfigures its learning algorithm with the new hyperparameters determined
by the memory. This mechanism, dubbed as Episodic Policy Gradient
Training (EPGT), enables an episodic learning process, and jointly
learns the policy and the learning algorithm's hyperparameters within
a single run. Experimental results on both continuous and discrete
environments demonstrate the advantage of using the proposed method
in boosting the performance of various policy gradient algorithms.
\end{abstract}
\global\long\def\sigmoid{sigmoid}%

\section{Introduction}

The current success of deep reinforcement learning relies on the ability
to use gradient-based optimizations for policy and value learning
\cite{mnih2015human,silver2017mastering}. Approaches such as \emph{policy
gradient} (PG) methods have achieved remarkable results in various
domains including games \cite{mnih2016asynchronous,schulman2017proximal,wu2017scalable,fujimoto2018addressing},
robotics \cite{kohl2004policy,peters2006policy} or even natural language
processing \cite{ziegler2019fine}. However, the excellent performance
of PG methods is heavily dependent on tuning the algorithms' hyperparameters
\cite{duan2016benchmarking,zhang2021importance}. Applying a PG method
to new environments often requires different hyperparameter settings
and thus retuning \cite{henderson2018deep}. The large amount of hyperparameters
severely prohibits machine learning practitioners from fully utilizing
PG methods in different reinforcement learning environments. 

As a result, there is a huge demand for automating hyperparameter
selection for policy gradient algorithms, and it remains a critical
part of the Automated Machine Learning (AutoML) movement \cite{hutter2019automated}.
Automatic hyperparameter tuning has been well explored for supervised
learning. Simple methods such as grid search and random search are
effective although computationally expensive \cite{bergstra2012random,larochelle2007empirical}.
Other complex methods such as Bayesian Optimization (BO \cite{snoek2012practical})
and Evolutionary Algorithms (EA \cite{fiszelew2007finding}) can efficiently
search for optimal hyperparameters. Yet, they still need multiple
training runs, have difficulty scaling to high-dimensional settings
\cite{rana2017high} or require extensive parallel computation \cite{jaderberg2017population}.
Recent attempts introduce online hyperparameter scheduling, that jointly
optimizes the hyperparameters and parameters in single run overcoming
local optimality of training with fixed hyperparameters and showing
great potential for supervised and reinforcement learning \cite{jaderberg2017population,xu2018meta,NEURIPS2019_743c41a9,parker2020provably}. 

However, one loophole remains. These approaches do not model the \emph{context
of training} in the optimization process, and the problem is often
treated as a stateless bandit or greedy optimization \cite{NEURIPS2019_743c41a9,parker2020provably}.
Ignoring the context prevents the use of episodic experiences that
can be critical in optimization and planning. As an example, we humans
often rely on past outcomes of our actions and their contexts to optimize
decisions (e.g. we may use past experiences of traffic to not return
home from work at 5pm). Episodic memory plays a major role in human
brains, facilitating recreation of the past and supporting decision
making via recall of episodic events \cite{tulving2002episodic}.
We are motivated to use such a mechanism in training wherein, for
instance, the hyperparameters that helped overcome a past local optimum
in the loss surface can be reused when the learning algorithm falls
into a similar local optimum. This is equivalent to optimizing hyperparameters
based on training contexts. Patterns of bad or good training states
previously explored can be reused, and we refer to this process as
selecting hyperparameters. To implement this mechanism we use episodic
memory. Compared to other learning methods, the use of episodic memory
is non-parametric, fast and sample-efficient, and quickly directs
the agents towards good behaviors \cite{lengyel2008hippocampal,kumaran2016learning,blundell2016model}. 

\begin{figure*}
\begin{centering}
\includegraphics[width=0.8\textwidth]{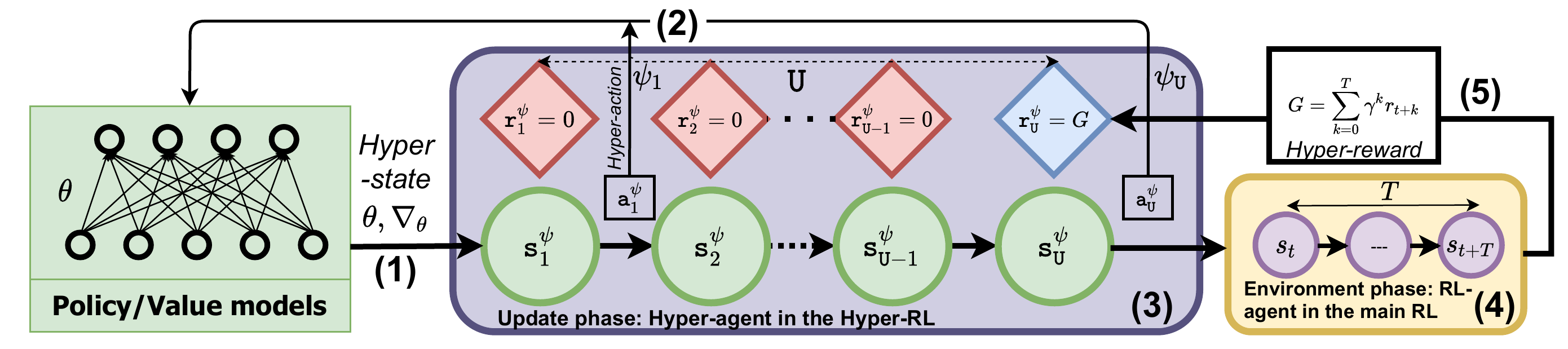}
\par\end{centering}
\caption{Hyper-RL structure. The hyper-state (green circle) is captured from
the PG models' parameters and gradients at every Hyper-RL step \textbf{(1)}.
Given the hyper-states, the hyper-agent takes hyper-actions, choosing
hyperparameters for the PG method to update the models \textbf{(2)}.
The update lasts $\mathtt{U}$ steps. After the last update step \textbf{(3)},
the RL agent starts environment phase with the current policy, collecting
an empirical return $G$ after $T$ environment steps \textbf{(4)}.
$G$ is used as the hyper-reward for the last policy update step (blue
diamond) \textbf{(5).} Other update steps (red diamond) are assigned
with hyper-reward $0$. \label{fig:Reward-structure-of}}
\end{figure*}
This problem of formulating methods that can take the training context
into consideration and using them as episodic experiences in optimizing
hyperparameters remains unsolved. The first challenge is to effectively
represent the training context of PG algorithms that often involve
a large number of neural network parameters. The second challenge
is sample-efficiency. Current performant hyperparameter searches \cite{jaderberg2017population,parker2020provably}
often necessitate parallel interactions with the environments, which
is expensive and not always feasible in real-world applications. Ideally,
hyperparameter search methods should not ask for additional observations
that the PG algorithms already collect. If so, it must be solved as
efficiently as possible to allow efficient training of PG algorithms. 

We address both these issues with a novel solution, namely Episodic
Policy Gradient Training (EPGT)--a PG training scheme that allows
on-the-fly hyperparameter optimization based on episodic experiences.
The idea is to formulate hyperparameter scheduling as a Markov Decision
Process (MDP), dubbed as Hyper-RL. In the Hyper-RL, an agent (hyper-agent)
acts to optimize hyperparameters for the PG algorithms that optimize
the policy for the agent of the main RL (RL-agent). The two agents
operate alternately: the hyper-agent acts to reconfigure the PG algorithms
with different hyperparameters, which ultimately changes the policy
of the RL agent (update phase); the RL agent then acts to collect
returns (environment phase), which serves as the rewards for the hyper-agent.
To build the Hyper-RL, we propose mechanisms to model its state, action
and reward. In particular, we model the training context as the state
of the Hyper-RL by using neural networks to compress the parameters
and gradients of PG models (policy/value networks) into low-dimensional
state vectors. The action in the Hyper-RL corresponds to the choice
of hyperparameters and the reward is derived from the RL agent's reward. 

We propose to solve the Hyper-RL through episodic memory. As an episodic
memory provides a direct binding from experiences (state-action) to
final outcome (return), it enables fast utilization of past experiences
and accelerates the searching of near-optimal policy \cite{lengyel2008hippocampal}.
Unlike other memory forms augmenting RL agents with stronger working
memory to cope with partial observations \cite{hung2019optimizing,pmlr-v119-le20b}
or contextual changes within an episode \cite{le2020neurocoder},
episodic memory persists across agent lifetime to maintain a global
value estimation. In our case, the memory estimates the value of a
state-action pair in the Hyper-RL by nearest neighbor memory lookup
\cite{pritzel2017neural}. To store learning experience, we use a
novel weighted average nearest neighbor writing rule that quickly
propagates the value inside the memory by updating multiple memory
slots per memory write. Our episodic memory is designed to cope with
noisy and sparse rewards in the Hyper-RL.

Our key contribution is to provide a new formulation for online hyperparameter
search leveraging context of previous training experiences, and demonstrate
that episodic memory is a feasible way to solve this. This is also
the first time episodic memory is designed for hyperparameter optimization.
Our rich set of experiments shows that EPGT works well with various
PG methods and diverse hyperparameter types, achieving higher rewards
without significant increase in computing resources. Our solution
has desirable properties, it is (i) computationally cheap and run
once without parallel computation, (ii) flexible to handle many hyperparameters
and PG methods, and (iii) shows consistent/significant performance
gains across environments and PG methods.

\section{Methods}

\subsection{Hyperparameter Reinforcement Learning (Hyper-RL) \label{subsec:Hyperparameter-Reinforcement-Lea}}

In this paper, we address the problem of online hyperparameter search.
We argue that in order to choose good values, hyperparameter search
(HS) methods should be aware of the past training states. This intuition
suggests that we should treat the HS problem as a standard MDP. Put
in the context of HS for RL, our HS algorithm becomes a Hyper-RL algorithm
besides the main RL algorithm. In Hyper-RL, the hyper-agent makes
decisions at each policy update step to configure the PG algorithm
with suitable hyperparameters $\psi$. The ultimate goal of the Hyper-RL
is the same as the main RL's: to maximize the return of the  RL agent. 

To construct the Hyper-RL, we define its state $\mathtt{s^{\psi}}$,
action $\mathtt{a}^{\psi}$ and reward $\mathtt{r^{\psi}}$. Hereafter,
we refer to them as hyper-state, hyper-action and hyper-reward to
avoid confusion with the main RL's $s$, $a$ and $r$. Fig. \ref{fig:Reward-structure-of}
illustrates the operation of Hyper-RL. In the update phase, the Hyper-RL
runs for $\mathtt{U}$ steps. At each step, taking the hyper-state
captured from the PG models' parameters and gradients, the hyper-agent
outputs hyper-actions, producing hyperparameters for PG algorithms
to update the policy/value networks accordingly. After the last update
(blue diamond), the resulting policy will be used by the RL agent
to perform the environment phase, collecting returns after $T$ environment
interactions. The returns will be used in the PG methods, and utilized
as hyper-reward for the last policy update step. Below we detail the
hyper-action, hyper-reward and hyper-state. 

\textbf{Hyper-action} A hyper-action $\mathtt{a}^{\psi}$ defines
the values for the hyperparameters $\psi$ of interest. For simplicity,
we assume the hyper-action is discrete by quantizing the range of
each hyperparameter into $\mathtt{B}$ discrete values. A hyper-action
$\mathtt{a^{\psi}}$ selects a set of discrete values, each of which
is assigned to a hyperparameter (see more Appendix A.2). 

\textbf{Hyper-reward} The hyper-reward $\mathtt{r}^{\psi}$ is computed
based on the empirical return that the RL agent collects in the environment
phase after hyperparameters are selected and used to update the policy.
The return is $G=\mathbb{E}_{s_{t:t+T},a_{t:t+T}}\left[\sum_{k=0}^{T}\gamma^{k}r_{t+k}\right]$
where $t$ and $T$ are the environment step and learning horizon,
respectively. Since there can be $\mathtt{\mathtt{U}}$ consecutive
policy update steps in the update phase, the last update step in the
update phase receives hyper-reward $G$ while others get zero hyper-reward,
making the Hyper-RL, in general, a sparse reward problem. That is,

\begin{equation}
\mathtt{r}_{i}^{\psi}=\begin{cases}
G & \textrm{if }i=\mathtt{U}\\
0 & \textrm{otherwise}
\end{cases}\label{eq:r_hyper}
\end{equation}
To define the objective for the Hyper-RL, we treat the update phase
as a learning episode. Each learning episode can lasts for multiple
of $\mathtt{U}$ update steps and for each step $i$ in the episode,
we aim to maximize the hyper-return $\mathtt{G}_{i}^{\psi}=\sum_{j\geq i}^{\mathtt{U}n}\mathtt{r}_{j}^{\psi}$
where $n\in\mathbb{N}^{+}$. In this paper, $n$ is simply set to
1 and thus, $\mathtt{G}_{i}^{\psi}=G$. 

\textbf{Hyper-state} A hyper-state $\mathtt{s^{\psi}}$ should capture
the current training state, which may include the status of the trained
model, the loss function or the amount of parameter update. We fully
capture $\mathtt{s^{\psi}}$ if we know exactly the loss surface and
the current value of the optimized parameters, which can result in
perfect hyperparameter choices. This, however, is infeasible in practice,
thus we only model observable features of the hyper-state space. The
Hyper-RL is then partially observable and noisy. In the following,
we propose a method to represent the hyper-state efficiently. 

\begin{algorithm}
\begin{algorithmic}[1]
\REQUIRE{A parametric policy function $\pi_\theta$ of the main RL algorithm $PG_{\psi}(\pi_{\theta},G)$ where $\psi$ is the set of hyperparamters for training $\pi_\theta$ and $G$ the empirical return collected by function  $Agent(\pi_\theta)$.}
\STATE{Initialize the episodic memory $\mathtt{M}=\emptyset$}
\FOR[loop over learning episodes]{$episode=1,2,...$}
\STATE{Initialize a buffer $\mathtt{D}=\emptyset$}\COMMENT{storing hyper-state, action, and reward within a learning episode}
\FOR[loop over policy updates]{$i=1,\dots\mathtt{U}$}

\STATE{Compute $\phi(\mathtt{s}^\psi_{i})$. Select  $\mathtt{a}^\psi_i$ by $\epsilon$-greedy with  $\mathtt{Q}\left(\mathtt{s}_i^\psi,\mathtt{a}^\psi\right)=\mathtt{M.read}
\left(\phi\left(\mathtt{s}^\psi_{i}\right),\mathtt{a}^\psi\right) $ (Eq. \ref{eq:mread})}
\STATE{Convert  $\mathtt{a}^\psi_i$ to the hyperparameter values $\psi_i$ and update $\theta \leftarrow PG_{\psi_i}(\pi_{\theta},G)$}
\STATE{Compute $\mathtt{r}^\psi_i$ (Eq. \ref{eq:r_hyper}). Add $(\phi(\mathtt{s}^\psi_{i}),\mathtt{a}^\psi_{i},\mathtt{r}^\psi_i)$ to $\mathtt{D}$}
\STATE{$\mathbf{if}$ $i$ == $\mathtt{U}$ $\mathbf{then}$ $G=Agent(\pi_\theta)$}
\ENDFOR
\STATE{Update episodic memory with $\mathtt{M.update(D)}$ (Eq. \ref{eq:knnw})}
\ENDFOR
\end{algorithmic}

\caption{Episodic Policy Gradient Training.\label{alg:Episodic-Policy-Optimizaiton.}}
\end{algorithm}
\textbf{Hyper-state representation}\label{subsec:Learning-state-modeling}
Our hypothesis is that one signature feature of the hyper-state is
the current value of optimized parameters $\theta$ and the derivatives
of the PG method's objective function w.r.t $\theta$. We maintain
a list of the last $N_{order}$ first-order derivatives: $\left\{ \nabla_{\theta n}\right\} _{n=1}^{N_{order}}$,
which preserves information of high-order derivatives (e.g. a second-order
derivative can be estimated by the difference between two consecutive
first-order derivatives). Let us denote the parameters and their derivatives,
often in tensor form, as $\theta=\left\{ W_{m}^{0}\right\} _{m=1}^{M}$
and $\nabla_{\theta n}=\left\{ W_{m}^{n}\right\} _{m=1}^{M}$ where
$M$ is the number of layers in the policy/value network. $\left\{ \theta,\nabla_{\theta n}\right\} $
can be denoted jointly $\left\{ W_{m}^{n}\right\} _{n=0,m=1}^{N_{order},M}$
or $\left\{ W_{m}^{n}\right\} $ for short (see Appendix B.4 for dimension
details of $W_{m}^{n}$).

Merely using $\left\{ W_{m}^{n}\right\} $ to represent the learning
state is still challenging since the number of parameters is enormous
as it often is in the case of recent PG methods. To make the hyper-state
tractable, we propose to use linear transformation to map the tensors
to lower-dimensional features and concatenate them to create the state
vector $\mathtt{s^{\psi}}=\left[s_{m}^{n}\right]_{n=0,m=1}^{N_{order},M}$.
Here, $s_{m}^{n}$ is the feature of $W_{m}^{n}$, computed as

\begin{equation}
s_{m}^{n}=\mathtt{vec}\left(W_{m}^{n}C_{m}^{n}\right)
\end{equation}
where $C_{m}^{n}\in\mathbb{R}^{d^{nm}\text{\ensuremath{\times d}}}$
is the transformation matrix, $d^{nm}$ the last dimension of $W_{m}^{n}$
($d^{nm}\gg d$) and $\mathtt{vec}\left(\cdot\right)$ the vectorize
operator, flattening the input tensor. To make our representation
robust, we propose to learn the transformation $C_{m}^{n}$ as described
in the next section. 

\textbf{Learning to represent hyper-state and memory key }We map $\mathtt{s^{\psi}}$
to its embedding by using a feed-forward neural network $\phi$, resulting
in the state embedding $\phi\left(\mathtt{s^{\psi}}\right)\in\mathbb{R}^{h}$.
$\phi\left(\mathtt{s}^{\psi}\right)$ later will be stored as the
key of the episodic memory. We can just use random $\phi$ and $C_{m}^{n}$
for simplicity. However, to encourage $\phi\left(\mathtt{s}^{\psi}\right)$
to store meaningful information of $\mathtt{s}^{\psi}$, we propose
to reconstruct $\mathtt{s}^{\psi}$ from $\phi\left(\mathtt{s}^{\psi}\right)$
via another decoder network $\omega$ and minimize the following reconstruction
error $\mathcal{L}_{rec}=\left\Vert \omega\left(\phi\left(\mathtt{s^{\psi}}\right)\right)-\mathtt{s^{\psi}}\right\Vert _{2}^{2}$.
Similar to \cite{blundell2016model}, we employ latent-variable probabilistic
models such as VAE to learn $C_{m}^{n}$ and update the encoder-decoder
networks. Thanks to using $C_{m}^{n}$ projection to lower dimensional
space, the hyper-state distribution becomes simpler and potential
for VAE reconstruction. Notably, the VAE is jointly trained online
with the RL agent and the episodic memory (more details in Appendix
A.3). 

\begin{figure*}
\begin{centering}
\includegraphics[width=1\textwidth]{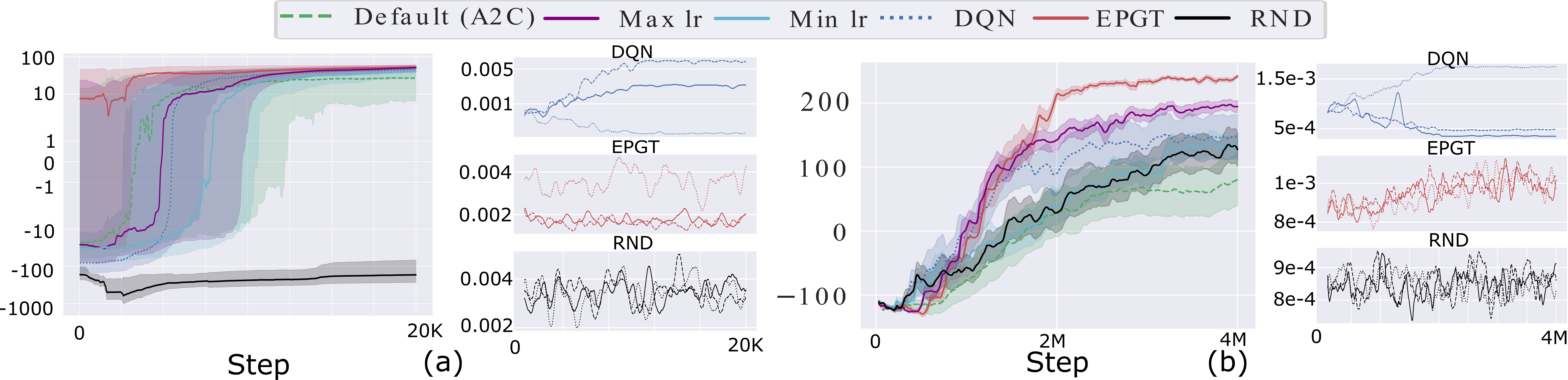}
\par\end{centering}
\caption{Performance on (a) MountainCarContinuous (log scale) and (b) BipedalWalker
over env. steps. In each plot, average return is on the left with
mean and std. over 10 runs. The right is smoothed (taking average
over a window of 100 steps) learning rate $\alpha$ found by the baselines
(first 3 runs). \label{fig:Training-performance-with}}
\end{figure*}

\subsection{Episodic Control for solving the Hyper-RL\label{subsec:Memory-operation}}

Theoretically, given the hyper-state, hyper-action and hyper-reward
clearly defined in the previous section, we can use any RL algorithm
to solve the Hyper-RL problem. However, in practice, the hyper-reward
is usually sparse and the number of steps of the Hyper-RL is usually
much smaller than that of the main RL algorithm ($\text{\ensuremath{\mathtt{U}}}\ll T$).
It means parametric methods (e.g. DQN) which require a huge number
of update steps are not suitable for learning a good approximation
of the Hyper-RL's Q-value function $\mathtt{Q}\left(\mathtt{s}_{i}^{\psi},\mathtt{a}_{i}^{\psi}\right)$. 

To quickly estimate $\mathtt{Q}\left(\mathtt{s}_{i}^{\psi},\mathtt{a}_{i}^{\psi}\right)$,
we maintain an episodic memory that lasts across learning episodes
and stores the outcomes of selecting hyperparameters from a given
hyper-state. We hypothesize that the training process involves hyper-states
that share similarities, which is suitable for episodic recall using
KNN memory lookup. Concretely, the episodic memory $\mathtt{M}$ binds
the learning experience $\left(\phi\left(\mathtt{s}_{i}^{\psi}\right),\mathtt{a}_{i}^{\psi}\right)$--the
key, where $\phi$ is an embedding function, to the approximated expected
hyper-return $\mathtt{\tilde{G}}_{i}^{\psi}$--the value. We index
the memory using key $\left(\phi\mathtt{\left(\mathtt{s}_{i}^{\psi}\right)},\mathtt{a}_{i}^{\psi}\right)$
to access the value, a.k.a $\mathtt{M}\left[\phi\mathtt{\left(\mathtt{s}_{i}^{\psi}\right)},\mathtt{a}_{i}^{\psi}\right]=\mathtt{\tilde{G}}_{i}^{\psi}$.
Computing and updating the $\mathtt{Q}\left(\mathtt{\mathtt{s}_{i}^{\psi}},\mathtt{a}_{i}^{\psi}\right)$
corresponds to two memory operators: $\mathtt{read}$ and $\mathtt{update}$.
The $\mathtt{read}\left(\phi\left(\mathtt{s}_{i}^{\psi}\right),\mathtt{a}_{i}^{\psi}\right)$
takes the hyper-state embedding plus hyper-action and returns the
hyper-state-action value $\mathtt{Q}\left(\mathtt{s}_{i}^{\psi},\mathtt{a}_{i}^{\psi}\right)$.
The $\mathtt{update}\left(\mathtt{D}\right)$ takes a buffer $\mathtt{D}$
containing observations $(\mathtt{s}_{i}^{\psi},\mathtt{a}_{i}^{\psi},\mathtt{r}_{i}^{\psi})_{i=1}^{\mathtt{U}}$,
and updates the content of the memory $\mathtt{M}$. The details of
the two operators are as follows. 

\textbf{Memory reading} Similarly to \cite{pritzel2017neural}, we
estimate the state-action value of any $\mathtt{s}_{i}^{\psi}$-$\mathtt{a}_{i}^{\psi}$
pair by:

\begin{eqnarray}
\mathtt{Q}\left(\mathtt{s}_{i}^{\psi},\mathtt{a}_{i}^{\psi}\right) & = & \mathtt{\mathtt{read}}\left(\mathtt{s}_{i}^{\psi},\mathtt{a}_{i}^{\psi}\right)\label{eq:mread}\\
 & = & \frac{\sum_{k=1}^{|\mathcal{N}(i)|}Sim\left(i,k\right)\mathtt{M}\left[\phi\left(\mathtt{s}_{k}^{\psi}\right),\mathtt{a}_{i}^{\psi}\right]}{\sum_{k=1}^{|\mathcal{N}(i)|}Sim\left(i,k\right)}\nonumber 
\end{eqnarray}
where $\mathcal{N}(i)$ denotes the neighbor set of the embedding
$\phi\left(\mathtt{s}_{i}^{\psi}\right)$ in $\mathtt{M}$ and $\phi\left(\mathtt{s}_{k}^{\psi}\right)$
the $k$-th nearest neighbor. $\mathcal{N}(i)$ includes $\phi\left(\mathtt{s}_{i}^{\psi}\right)$
if it exists in $\mathtt{M}$. $Sim\left(i,k\right)$ is a kernel
measuring the similarity between $\phi\left(\mathtt{s}_{k}^{\psi}\right)$
and $\phi\left(\mathtt{s}_{i}^{\psi}\right)$.

\textbf{Memory update} To cope with noisy observations from the Hyper-RL,
we propose to use weighted average to write the hyper-return to the
memory slots. Unlike max writing rule \cite{blundell2016model} that
always stores the best return, our writing propagates the average
return inside the memory, which helps cancel out the noise of the
Hyper-RL. In particular, for each observed transition in a learning
episode (stored in the buffer $\mathtt{D}$), we compute the hyper-return
$\mathtt{G}_{i}^{\psi}$. The hyper-return is then used to update
the memory such that the action value of $\phi\left(\mathtt{s}_{i}^{\psi}\right)$'s
neighbors is adjusted towards $\mathtt{G}_{i}^{\psi}$ with speeds
relative to the distances \cite{le2021modelbased}: 

\begin{align}
\mathtt{M}\left[\phi\left(\mathtt{s}_{k}^{\psi}\right),\mathtt{a}_{i}^{\psi}\right] & \leftarrow\mathtt{M}\left[\phi\left(\mathtt{s}_{k}^{\psi}\right),\mathtt{a}_{i}^{\psi}\right]+\beta\frac{\Delta_{ik}Sim\left(i,k\right)}{\sum_{k=1}^{|\mathcal{N}(i)|}Sim\left(i,k\right)}\label{eq:knnw}
\end{align}
where $\phi\left(\mathtt{s}_{k}^{\psi}\right)$ is the $k$-th nearest
neighbor of $\phi\left(\mathtt{s}_{i}^{\psi}\right)$ in $\mathcal{N}(i)$,
$\Delta_{ik}=\mathtt{G}_{i}^{\psi}-\mathtt{M}\left[\phi\left(\mathtt{s}_{k}^{\psi}\right),\mathtt{a}_{i}^{\psi}\right]$,
and $0<\beta<1$ the writing rate. If the key $\left(\phi\left(\mathtt{s}_{i}^{\psi}\right),\mathtt{a}_{i}^{\psi}\right)$
is not in $\mathtt{M}$, we also add $\left(\phi\left(\mathtt{s}_{i}^{\psi}\right),\mathtt{a}_{i}^{\psi},\mathtt{G}_{i}^{\psi}\right)$
to the memory. When the stored tuples exceed memory capacity $N_{mem}$,
the earliest added tuple will be removed. 

Under this formulation, $\mathtt{M}\left[\phi\left(\mathtt{s}_{i}^{\psi}\right),\mathtt{a}_{i}^{\psi}\right]$
is an approximation of the expected hyper-return collected by taking
the hyper-action $\mathtt{a}_{i}^{\psi}$ at the hyper-state $\mathtt{s}_{i}^{\psi}$
(see Appendix C for proof). As we update several neighbors at one
write, the hyper-return propagation inside the episodic memory is
faster and helps to handle the sparsity of the Hyper-RL. Unless stated
otherwise, we use the same neighbor size $|\mathcal{N}(i)|$ for both
reading and writing process, denoted as $K$ for short. 

\textbf{Integration with PG methods} Our episodic control mechanisms
can be used to estimate the hyper-state-action-value of the Hyper-RL.
The hyper-agent uses that value to select the hyper-action through
$\epsilon$-greedy policy and schedule the hyperparameters of PG methods.
Algo. \ref{alg:Episodic-Policy-Optimizaiton.}, Episodic Policy Gradient
Training (EPGT), depicts the use of our episodic control with a generic
PG method.

\section{Experimental results}

Across experiments, we examine EPGT with different PG methods including
A2C \cite{mnih2016asynchronous}, ACKTR \cite{wu2017scalable} and
PPO \cite{schulman2017proximal}. We benchmark EPGT against the original
PG methods with tuned hyperparameters and 4 recent hyperparameter
search methods. The experimental details can be found in the Appendix
B. 

\subsection{Why episodic control?}

In this section, we validate the choice of episodic control to solve
the proposed Hyper-RL problem. As such, we choose A2C as the PG method
and examine EPGT, random hyper-action (RND) and DQN \cite{mnih2015human}
as 3 methods to schedule the learning rate ($\alpha$) for A2C. We
also compare with A2C using different fixed-$\alpha$ within the search
range (default, min and max learning rates). We test on 2 environments:
Mountain Car Continuous (MCC) and Bipedal Walker (BW) with long and
short learning rate search ranges ($[4\times10^{-5},10^{-2}]$ and
$[2.8\times10^{-4},1.8\times10^{-3}]$, respectively).

Fig. \ref{fig:Training-performance-with} demonstrates the learning
curves and learning rate schedules found by EPGT, RND and DQN. In
MCC, the search range is long, which makes RND performance unstable,
far lower than the fixed-$\alpha$ A2Cs. DQN also struggles to learn
good $\alpha$ schedule for A2C since the number of trained environment
steps is only 20,000, which corresponds to only 4,000 steps in the
Hyper-RL. This might not be enough to train DQN's value network and
leads to slower learning. On the contrary, EPGT helps A2C achieve
the best performance faster than any other baseline. In BW, thanks
to shorter search range and large number of training steps, RND and
DQN show better results, yet still underperform the best fixed-$\alpha$
A2C. By contrast, EPGT outperforms the best fixed-$\alpha$ A2C by
a significant margin, which confirms the benefit of episodic dynamic
hyperparameter scheduling.

Besides performance plots, we visualize the selected values of learning
rates over training steps for the first 3 runs of each baseline. Interestingly,
DQN finds more consistent values, often converging to extreme learning
rates, indicating that the DQN mostly selects the same action for
any state, which is unreasonable. EPGT, on the other hand, prefers
moderate learning rates, which keep changing depending on the state.
Compared to random schedules by RND, those found by EPGT have a pattern,
either gradually decreasing (MCC) or increasing (BW). In terms of
running time, EPGT runs slightly slower than A2C without any scheduler,
yet much faster than DQN (see Appendix's Table \ref{tab:Computing-cost-of}).

\begin{table}
\begin{centering}
\begin{tabular}{ccccc}
\hline 
{\small{}Model} & {\small{}HalfCheetah} & {\small{}Hopper} & {\small{}Ant} & {\small{}Walker}\tabularnewline
\hline 
{\small{}TMG$^{\spadesuit}$} & {\small{}1,568} & {\small{}378} & {\small{}950} & {\small{}492}\tabularnewline
{\small{}HOOF$^{\spadesuit}$} & {\small{}1,523} & {\small{}350} & {\small{}952} & {\small{}467}\tabularnewline
{\small{}HOOF$^{\diamondsuit}$} & {\small{}1,427\textpm 293} & {\small{}452\textpm 40.7} & {\small{}954\textpm 8.57} & {\small{}674\textpm 195}\tabularnewline
\hline 
{\small{}EPGT} & \textbf{\small{}2,530}{\small{}\textpm }\textbf{\small{}1268} & \textbf{\small{}603}{\small{}\textpm }\textbf{\small{}187} & {\small{}1,083\textpm 126} & \textbf{\small{}888}{\small{}\textpm }\textbf{\small{}425}\tabularnewline
\hline 
\end{tabular}
\par\end{centering}
\caption{EPGT vs sequential online hyperparameter search (A2C as the PG). Bold
denotes statistically better results in terms of Cohen effect size
$>0.5$. We train agents for 5 million steps and report the mean (and
std. if applicable) over 10 runs. $\spadesuit$ is from \cite{NEURIPS2019_743c41a9}
(no std. reported) and $\diamondsuit$ is our run.\label{tab:EPO-vs-sequential}}
\end{table}
\begin{table}
\noindent \begin{centering}
\begin{tabular}{ccccc}
\hline 
\multirow{1}{*}{{\small{}Model}} & {\small{}BW} & {\small{}LLC} & {\small{}Hopper} & {\small{}IDP}\tabularnewline
\hline 
{\small{}PBT$^{\lyxmathsym{\textdegree}}$} & {\small{}223} & {\small{}159} & {\small{}1492} & {\small{}8,893}\tabularnewline
{\small{}PB2$^{\lyxmathsym{\textdegree}}$} & {\small{}276} & \textbf{\small{}235} & {\small{}2,346} & {\small{}8,893}\tabularnewline
{\small{}PB2$^{\diamondsuit}$} & {\small{}280} & {\small{}223} & {\small{}2,156} & {\small{}9,253}\tabularnewline
\hline 
{\small{}EPGT} & \textbf{\small{}282} & \textbf{\small{}235} & \textbf{\small{}3,253} & \textbf{\small{}9,322}\tabularnewline
\hline 
\end{tabular}
\par\end{centering}
\caption{EPGT vs parallel online hyperparameter search (PPO as the PG). Following
\cite{parker2020provably}, we train the PPO agents for 1 million
steps and report the best median over 10 runs.$\lyxmathsym{\protect\textdegree}$
denotes the numbers reported in \cite{parker2020provably}, and $\diamondsuit$
is our run.\label{tab:EPO-vs-parallel}}
\end{table}

\subsection{EPGT vs online hyperparameter search methods\label{subsec:EPO-vs-other}}

Our main baselines are existing methods for dynamic tuning of hyperparameters
of policy gradient algorithms, which can be divided into 2 groups:
(i) sequential HOOF \cite{NEURIPS2019_743c41a9} and Meta-gradient
\cite{xu2018meta} and (ii) parallel PBT \cite{jaderberg2017population}
and PB2 \cite{parker2020provably}. We follow the same experimental
setting (PG configuration and environment version) and apply our EPGT
to the same set of optimized hyperparameters, keeping other hyperparameters
as in other baselines. We also rerun the baselines HOOF and PB2 using
our codebase to ensure fair comparison. For the first group, the PG
method is A2C and only the learning rate is optimized, while for the
second group, the PG method is PPO and we optimize 4 hyperparameters
(learning rate $\alpha$, batch size $b$, GAE $\lambda$ and PPO
clip $\epsilon$).

Table \ref{tab:EPO-vs-sequential} reports the mean test performance
of EPGT against Tuned Meta-gradient (TMG) and HOOF on 4 Mujoco environments.
EPGT demonstrates better results in all 4 tasks where HalfCheetah,
Hopper and Walker observe significant gain. Notably, compared to HOOF,
EPGT exhibits higher mean and variance, indicating that EPGT can find
distinctive solutions, breaking the local optimum bottleneck of other
baselines. 

Table \ref{tab:EPO-vs-parallel} compares EPGT with PBT and PB2 on
corresponding environments and evaluation metrics. In the four tasks
used in PB2 paper, EPGT achieves better median best score for 3 tasks
while maintaining competitive performance in LLC task. We note that
EPGT is jointly trained with the PG methods in a single run and thus,
achieves this excellent performance without parallel interactions
with the environments as PB2 or PBT. Learning curves of our runs for
the above tasks are in Appendix B.3.

\subsection{EPGT vs grid-search/manual tuning\label{subsec:EPO-vs-tuned}}

\textbf{Atari} We now examine EPGT on incremental sets of hyperparameters.
We adopt 6 standard Atari games and train 2 PG methods: ACKTR and
PPO for 20 million steps per game. For ACKTR, we apply EPGT to schedule
the trust region radius $\delta$, step size $\eta$ and the value
loss coefficient $l_{v}$. For PPO, the optimized hyperparameters
are learning rate $\alpha$, trust region clip $\epsilon$ and batch
size $b$. These are important and already tuned hyperparameters for
Atari task by prior works. We form 3 hyperparameter sets for each
PG method. For each set, we further perform grid search near the default
hyperparameters and record the best tuned results. We compare these
results with EPGT's and report the relative improvement on human normalized
score (see Fig. \ref{fig:Performance-improvement-on} for the case
of ACKTR, full in Appendix Fig. \ref{fig:Atari-games:-Average-1-1}).

The results indicate that, for all hyperparameter sets, EPGT on average
show gains up to more than 10\% over tuned PG methods. For certain
games, the performance gain can be more than 30\%. Jointly optimizing
more hyperparameters is generally better for PPO while optimizing
only $\delta$ gets the most improvement for ACKTR. 

\textbf{Mujoco }Here, we conduct experiments on 6 Mujoco environments:
HalfCheetah, Hooper, Walker2d, Swimmer, Ant and Humanoid. For the
last two challenging tasks, we train with 10M steps while the others
1M steps. The set of optimized hyperparameters are $\left\{ \alpha,\epsilon,\lambda,b\right\} $.
The baseline Default (PPO) has fixed hyperparameters, which are well-tuned
by previous works, and PB2 uses the same hyperparameter search range
as our method. Random hyper-action (RND) baseline is included to see
the difference between random and episodic policy in Hyper-RL formulation. 

On 6 Mujoco tasks, on average, EPGT helps PPO earn more than 583 score
while PB2 fails to clearly outperform the tuned PPO (see more in Appendix
Fig. \ref{fig:Mujoco-games:-Average}). Fig. \ref{fig:Performance-on-2}
(left) illustrates the result on HalfCheetah where performance gap
between EPGT and other baselines can be clearly seen. Despite using
the same search range, PB2 and RND show lower average return. We include
the hyperparameters used by EPGT and RND throughout training in Fig.
\ref{fig:Performance-on-2} (right). Overall, EPGT's schedules do
not diverge much from the default values, which are already well-tuned.
However, we can see a pattern of using smaller hyperparameters during
middle phase of training, which aligns with the moments when there
are changes in the performance. 

\begin{figure}
\begin{centering}
\includegraphics[width=0.95\columnwidth]{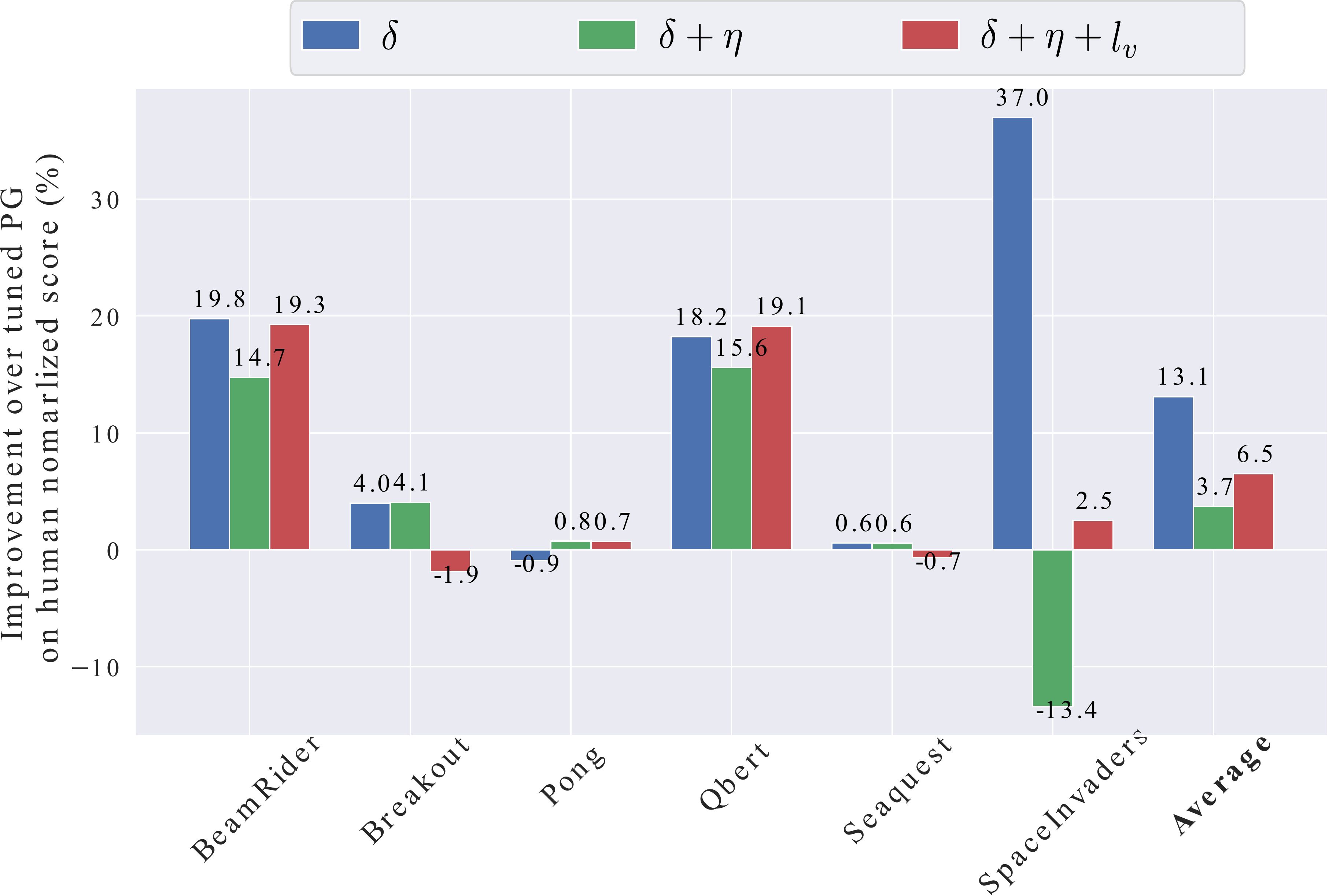}
\par\end{centering}
\caption{Performance improvement on 6 Atari games applying EPGT to optimize
incremental sets of hyperparameters with ACKTR as PG. We run each
game for 20 million frames and report the average over 5 runs.\label{fig:Performance-improvement-on}}
\end{figure}
\begin{figure}
\begin{centering}
\includegraphics[width=0.95\columnwidth]{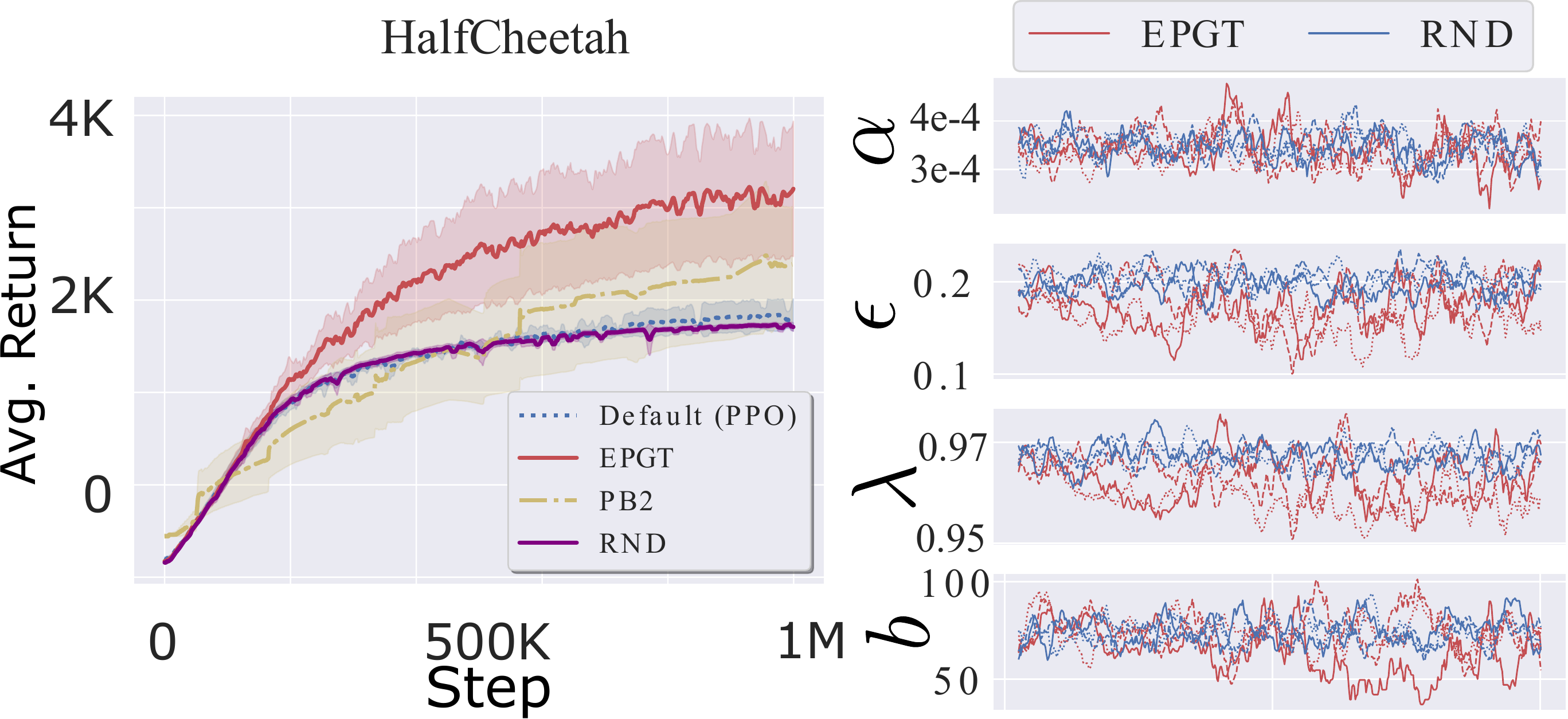}
\par\end{centering}
\caption{Performance on the representative HalfCheetah task over env. steps.
The left is testing return over training iterations (mean $\pm$ std.
over 10 runs) and the right hyperparameters schedule for PG methods
found by EPGT and RND in the first 3 runs. \label{fig:Performance-on-2}}
\end{figure}
\begin{figure*}
\begin{centering}
\includegraphics[width=0.95\textwidth]{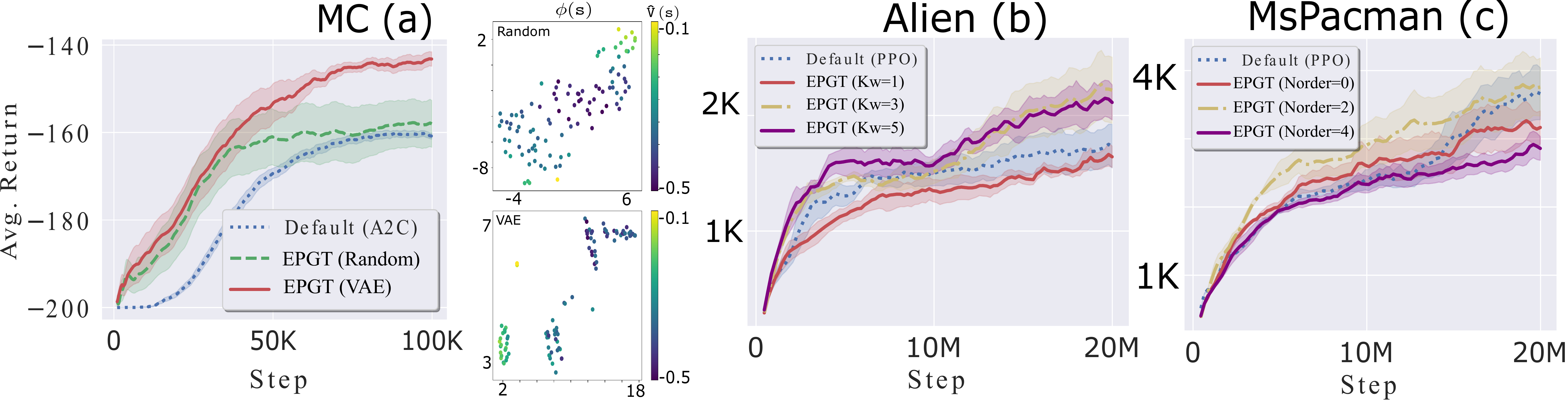}
\par\end{centering}
\caption{(a) Performance (left) and hyper-state representations $\phi\left(\mathtt{s}^{\psi}\right)$
(right) on Mountain Car (MC) using PG A2C where t-SNE is used to project
$\phi\left(\mathtt{s}^{\psi}\right)$ to 2d space, showing the quality
of representation learned by VAE (below) versus Random mapping (above).
Performance on Alien (b) and MsPacman (c) using PG PPO with different
$K_{w}$ and $N_{order}$, respectively. The curves are mean and std.
over 5 runs.\label{fig:(a)-Learning-state}}
\end{figure*}

\subsection{Ablation studies }

In this section, we describe the hyperparameter selection for EPGT
used in above the experiments. We note that although EPGT introduces
several hyperparameters, it is efficient to pick reasonable values
and keep using them across tasks. 

\textbf{Learning to represent the hyper-state} The hyper-state is
captured by projecting the model's weights and their gradients to
a low-dimensional vectors using $C_{m}^{n}$. The state is further
transformed to the memory's key using the mapping network $\phi$.
Here, we validate the choice of using VAE to learn $C_{m}^{n}$ and
$\phi$ by comparing it with random mapping. We use PG A2C and test
the two EPGT variants on Mountain Car (MC). Fig. \ref{fig:(a)-Learning-state}
(a, left) demonstrates that EPGT with VAE training learns fastest
and achieves the best convergence. EPGT with random projections can
learn fast but shows similar convergence as the original A2C. 

We visualize the final representations $\phi\left(\mathtt{s}^{\psi}\right)$
by using t-SNE and use colors to denote the corresponding average
values $\mathtt{\hat{V}\left(\mathtt{s}^{\psi}\right)}=\sum_{a}\mathtt{M\left(\phi\left(\mathtt{s^{\psi}}\right),a^{\psi}\right)}$
in Fig. \ref{fig:(a)-Learning-state} (a, right). The upper figure
is randomly projected hyper-states and the lower VAE-trained ones
at 5,000 environment step. From both figures, we can see that similar-value
states tend to lie together, which validates the hypothesis on existing
similar training contexts. Compared to the random ones, the representations
learned by VAE exhibit clearer clusters. Cluster separation is critical
for nearest neighbor memory access in episodic control, and thus explains
why VAE-trained EPGT outperforms random EPGT significantly. Notably,
training the VAE is inexpensive. Empirical results demonstrates that
with reasonable hyper-state sizes, the VAE converges quickly (see
Appendix Fig. \ref{fig:Reconstruction-loss-training}).

\textbf{Writing rule} To verify the contribution of our proposed writing
rule, we test different number of writing neighbor size ($K_{w}$).
In this experiment, the reading size is fixed to 3 and different from
the writing size. Fig. \ref{fig:(a)-Learning-state} (b) shows the
learning curves of EPGT using different $K_{w}$ against the original
PPO. When $K_{w}=1$, our rule becomes single-slot writing as in \cite{blundell2016model,pritzel2017neural},
which even underperforms using default hyperparameters. By contrast,
increasing $K_{w}=3$ boosts EPGT's performance dramatically, on average
improving PPO by around 500 score in Alien game. Increasing $K_{w}$
further seems not helpful since it may create noise in writing. Thus,
we use $K_{w}=3$ in all of our experiments. Others showing our average
writing rule is better than traditional max rule and examining different
numbers of general neighbor size $K$ are in Appendix B.4. 

\textbf{Order of representation} Finally, we examine EPGT's performance
with different order of representation ($N_{order}$). $N_{order}=0$
means the hyper-state only includes the parameters $\theta$. Increasing
$N_{order}$ gives more information, providing better state representations.
That holds true in MsPacman game when we increase the order from $0$
to $2$ as shown in Fig. \ref{fig:(a)-Learning-state} (c). However,
when $N_{order}$ is set to 4, the performance drops since the hyper-states
now are in a very high dimension ($16K$) and VAE does not work well
in this case. Hence, we use $N_{order}=2$ for all of our experiments.

\section{Related works}

\textbf{Hyperparameter} \textbf{search} Automatic hyperparameter tuning
generally requires multiple training runs. Parallel search methods
such as grid or random search \cite{bergstra2012random,larochelle2007empirical}
perform multiple runs concurrently and pick the hyperparameters that
achieve best result. These methods are simple yet expensive. Sequential
search approaches reduce the number of runs by consecutively executing
experiments using a set of candidate hyperparameters and utilize the
evaluation result to guide the subsequent choice of candidates \cite{hutter2011sequential}.
Bayesian Optimization approaches \cite{brochu2010tutorial} exploit
the previous experimental results to update the posterior of a Bayesian
model of hyperparameters and explore promising hyperparameter regions.
They have been widely used in hyperparameter tuning for various machine
learning algorithms including deep learning \cite{snoek2012practical,klein2017fast}.
Recently, to speed up the process, distributed versions of BO are
also introduced to evaluate in parallel batches of hyperparameter
settings \cite{gonzalez2016batch,chen2018bayesian}. 

However, these approaches still suffer from the issue of computational
inefficiency, demanding high computing resources, increasing the training
time significantly. If applied to RL, they require more environment
interactions, which leads to sample inefficiency. In addition, the
hyperparameters found by these methods are usually fixed, which can
be suboptimal \cite{luketina2016scalable}. Inspired by biological
evolution, population-based methods initially start as random search
then select best performing hyperparameter instances to generate subsequent
hyperparameter candidates, providing a hybrid solution between parallel
and sequential search \cite{fiszelew2007finding,young2015optimizing}.

Recent works propose using evolutionary algorithms to jointly learn
the weights and hyperparameters of neural networks under supervised
training \cite{jaderberg2017population,li2019generalized}. In BPT
\cite{jaderberg2017population} as an example, multiple training are
executed asynchronously and evaluated periodically. Under-performing
models are replaced by better ones whose hyperparameters evolve to
explore better configurations. This approach allows hyperparameter
scheduling on-the-fly but still requires a large number of parallel
runs and are thus unsuitable for machines with small computational
budget. 

\textbf{On-the-fly hyperparameter search for reinforcement learning}
Early works on gradient-based hyperparameter search focus on learning
rate adjustment \cite{sutton1992adapting,bengio2000gradient}. The
approach has been recently extended to RL by using the meta-gradient
of the return function to adjust the hyperparameters such as discount
factor or bootstrapping parameter \cite{xu2018meta}. Hence, in this
approach, the return needs to be a differentiable function w.r.t the
hyperparameters, which cannot extend to any hyperparameter type such
as ``clip'' or policy gradient algorithm such as TRPO.

HOOF \cite{NEURIPS2019_743c41a9} is an alternative to meta-gradient
methods wherein hyperparameter optimization is done via random search
and weighted important sampling. The method relies on off-policy estimate
of the value of the policy, which is known to have high variance and
thus requires enforcing additional KL constraint. The search is also
limited to some specific hyperparameters. Population-based approaches
have been applied to RL hyperparameter search. These methods become
more efficient by utilizing off-policy PG's samples \cite{tang2020online}
and small-size population \cite{parker2020provably}, showing better
results than PBT or BO in RL domains. However, they still suffer from
the inherited expensive computation issue of population-based training.
All of these prior works do not formulate hyperparameter search as
a MDP, bypassing the context of training, which is addressed in this
paper.

\section{Discussion}

We introduced Episodic Policy Gradient Training (EPGT), a new approach
for online hyperparameter search using episodic memory. Unlike prior
works, EPGT formulates the problem as a Hyper-RL and focuses on modeling
the training state to utilize episodic experiences. Then, an episodic
control with improved writing mechanisms is employed to search for
optimal hyperparameters on-the-fly. Our experiments demonstrate that
EPGT can augment various PG algorithms to optimize different types
of hyperparameters, achieving better results. Current limitations
of EPGT are the coarse discrete action spaces and simplified hyper-state
modeling using linear mapping. We will address these issues and extend
our approach to supervised training in future works.

\subsubsection*{ACKNOWLEDGMENTS}

This research was partially funded by the Australian Government through
the Australian Research Council (ARC). Prof Venkatesh is the recipient
of an ARC Australian Laureate Fellowship (FL170100006).

\bibliographystyle{plain}
\bibliography{epo}

\begin{thebibliography}{43}
\providecommand{\natexlab}[1]{#1}

\bibitem[{Bengio(2000)}]{bengio2000gradient}
Bengio, Y. 2000.
\newblock Gradient-based optimization of hyperparameters.
\newblock \emph{Neural computation}, 12(8): 1889--1900.

\bibitem[{Bergstra and Bengio(2012)}]{bergstra2012random}
Bergstra, J.; and Bengio, Y. 2012.
\newblock Random search for hyper-parameter optimization.
\newblock \emph{Journal of machine learning research}, 13(2).

\bibitem[{Blundell et~al.(2016)Blundell, Uria, Pritzel, Li, Ruderman, Leibo,
  Rae, Wierstra, and Hassabis}]{blundell2016model}
Blundell, C.; Uria, B.; Pritzel, A.; Li, Y.; Ruderman, A.; Leibo, J.~Z.; Rae,
  J.; Wierstra, D.; and Hassabis, D. 2016.
\newblock Model-free episodic control.
\newblock \emph{arXiv preprint arXiv:1606.04460}.

\bibitem[{Brochu, Cora, and De~Freitas(2010)}]{brochu2010tutorial}
Brochu, E.; Cora, V.~M.; and De~Freitas, N. 2010.
\newblock A tutorial on Bayesian optimization of expensive cost functions, with
  application to active user modeling and hierarchical reinforcement learning.
\newblock \emph{arXiv preprint arXiv:1012.2599}.

\bibitem[{Chen et~al.(2018)Chen, Huang, Wang, Antonoglou, Schrittwieser,
  Silver, and de~Freitas}]{chen2018bayesian}
Chen, Y.; Huang, A.; Wang, Z.; Antonoglou, I.; Schrittwieser, J.; Silver, D.;
  and de~Freitas, N. 2018.
\newblock Bayesian optimization in alphago.
\newblock \emph{arXiv preprint arXiv:1812.06855}.

\bibitem[{Duan et~al.(2016)Duan, Chen, Houthooft, Schulman, and
  Abbeel}]{duan2016benchmarking}
Duan, Y.; Chen, X.; Houthooft, R.; Schulman, J.; and Abbeel, P. 2016.
\newblock Benchmarking deep reinforcement learning for continuous control.
\newblock In \emph{International conference on machine learning}, 1329--1338.
  PMLR.

\bibitem[{Fiszelew et~al.(2007)Fiszelew, Britos, Ochoa, Merlino, Fern{\'a}ndez,
  and Garc{\'\i}a-Mart{\'\i}nez}]{fiszelew2007finding}
Fiszelew, A.; Britos, P.; Ochoa, A.; Merlino, H.; Fern{\'a}ndez, E.; and
  Garc{\'\i}a-Mart{\'\i}nez, R. 2007.
\newblock Finding optimal neural network architecture using genetic algorithms.
\newblock \emph{Advances in computer science and engineering research in
  computing science}, 27: 15--24.

\bibitem[{Fujimoto, Hoof, and Meger(2018)}]{fujimoto2018addressing}
Fujimoto, S.; Hoof, H.; and Meger, D. 2018.
\newblock Addressing function approximation error in actor-critic methods.
\newblock In \emph{International Conference on Machine Learning}, 1587--1596.
  PMLR.

\bibitem[{Gonz{\'a}lez et~al.(2016)Gonz{\'a}lez, Dai, Hennig, and
  Lawrence}]{gonzalez2016batch}
Gonz{\'a}lez, J.; Dai, Z.; Hennig, P.; and Lawrence, N. 2016.
\newblock Batch Bayesian optimization via local penalization.
\newblock In \emph{Artificial intelligence and statistics}, 648--657. PMLR.

\bibitem[{Henderson et~al.(2018)Henderson, Islam, Bachman, Pineau, Precup, and
  Meger}]{henderson2018deep}
Henderson, P.; Islam, R.; Bachman, P.; Pineau, J.; Precup, D.; and Meger, D.
  2018.
\newblock Deep reinforcement learning that matters.
\newblock In \emph{Proceedings of the AAAI Conference on Artificial
  Intelligence}, volume~32.

\bibitem[{Hung et~al.(2019)Hung, Lillicrap, Abramson, Wu, Mirza, Carnevale,
  Ahuja, and Wayne}]{hung2019optimizing}
Hung, C.-C.; Lillicrap, T.; Abramson, J.; Wu, Y.; Mirza, M.; Carnevale, F.;
  Ahuja, A.; and Wayne, G. 2019.
\newblock Optimizing agent behavior over long time scales by transporting
  value.
\newblock \emph{Nature communications}, 10(1): 1--12.

\bibitem[{Hutter, Hoos, and Leyton-Brown(2011)}]{hutter2011sequential}
Hutter, F.; Hoos, H.~H.; and Leyton-Brown, K. 2011.
\newblock Sequential model-based optimization for general algorithm
  configuration.
\newblock In \emph{International conference on learning and intelligent
  optimization}, 507--523. Springer.

\bibitem[{Hutter, Kotthoff, and Vanschoren(2019)}]{hutter2019automated}
Hutter, F.; Kotthoff, L.; and Vanschoren, J. 2019.
\newblock \emph{Automated machine learning: methods, systems, challenges}.
\newblock Springer Nature.

\bibitem[{Jaderberg et~al.(2017)Jaderberg, Dalibard, Osindero, Czarnecki,
  Donahue, Razavi, Vinyals, Green, Dunning, Simonyan
  et~al.}]{jaderberg2017population}
Jaderberg, M.; Dalibard, V.; Osindero, S.; Czarnecki, W.~M.; Donahue, J.;
  Razavi, A.; Vinyals, O.; Green, T.; Dunning, I.; Simonyan, K.; et~al. 2017.
\newblock Population based training of neural networks.
\newblock \emph{arXiv preprint arXiv:1711.09846}.

\bibitem[{Klein et~al.(2017)Klein, Falkner, Bartels, Hennig, and
  Hutter}]{klein2017fast}
Klein, A.; Falkner, S.; Bartels, S.; Hennig, P.; and Hutter, F. 2017.
\newblock Fast bayesian optimization of machine learning hyperparameters on
  large datasets.
\newblock In \emph{Artificial Intelligence and Statistics}, 528--536. PMLR.

\bibitem[{Kohl and Stone(2004)}]{kohl2004policy}
Kohl, N.; and Stone, P. 2004.
\newblock Policy gradient reinforcement learning for fast quadrupedal
  locomotion.
\newblock In \emph{IEEE International Conference on Robotics and Automation,
  2004. Proceedings. ICRA'04. 2004}, volume~3, 2619--2624. IEEE.

\bibitem[{Kumaran, Hassabis, and McClelland(2016)}]{kumaran2016learning}
Kumaran, D.; Hassabis, D.; and McClelland, J.~L. 2016.
\newblock What learning systems do intelligent agents need? Complementary
  learning systems theory updated.
\newblock \emph{Trends in cognitive sciences}, 20(7): 512--534.

\bibitem[{Larochelle et~al.(2007)Larochelle, Erhan, Courville, Bergstra, and
  Bengio}]{larochelle2007empirical}
Larochelle, H.; Erhan, D.; Courville, A.; Bergstra, J.; and Bengio, Y. 2007.
\newblock An empirical evaluation of deep architectures on problems with many
  factors of variation.
\newblock In \emph{Proceedings of the 24th international conference on Machine
  learning}, 473--480.

\bibitem[{Le et~al.(2021)Le, George, Abdolshah, Tran, and
  Venkatesh}]{le2021modelbased}
Le, H.; George, T.~K.; Abdolshah, M.; Tran, T.; and Venkatesh, S. 2021.
\newblock Model-Based Episodic Memory Induces Dynamic Hybrid Controls.
\newblock In \emph{Thirty-Fifth Conference on Neural Information Processing
  Systems}.

\bibitem[{Le, Tran, and Venkatesh(2019)}]{le2019learning}
Le, H.; Tran, T.; and Venkatesh, S. 2019.
\newblock Learning to remember more with less memorization.
\newblock In \emph{Proceedings of the 7th International Conference on Learning
  Representations}.

\bibitem[{Le, Tran, and Venkatesh(2020)}]{pmlr-v119-le20b}
Le, H.; Tran, T.; and Venkatesh, S. 2020.
\newblock Self-Attentive Associative Memory.
\newblock In \emph{Proceedings of the 37th International Conference on Machine
  Learning}.

\bibitem[{Le and Venkatesh(2020)}]{le2020neurocoder}
Le, H.; and Venkatesh, S. 2020.
\newblock Neurocoder: Learning General-Purpose Computation Using Stored Neural
  Programs.
\newblock \emph{arXiv preprint arXiv:2009.11443}.

\bibitem[{Lengyel and Dayan(2008)}]{lengyel2008hippocampal}
Lengyel, M.; and Dayan, P. 2008.
\newblock Hippocampal contributions to control: the third way.
\newblock In \emph{Advances in neural information processing systems},
  889--896.

\bibitem[{Li et~al.(2019)Li, Spyra, Perel, Dalibard, Jaderberg, Gu, Budden,
  Harley, and Gupta}]{li2019generalized}
Li, A.; Spyra, O.; Perel, S.; Dalibard, V.; Jaderberg, M.; Gu, C.; Budden, D.;
  Harley, T.; and Gupta, P. 2019.
\newblock A generalized framework for population based training.
\newblock In \emph{Proceedings of the 25th ACM SIGKDD International Conference
  on Knowledge Discovery \& Data Mining}, 1791--1799.

\bibitem[{Luketina et~al.(2016)Luketina, Berglund, Greff, and
  Raiko}]{luketina2016scalable}
Luketina, J.; Berglund, M.; Greff, K.; and Raiko, T. 2016.
\newblock Scalable gradient-based tuning of continuous regularization
  hyperparameters.
\newblock In \emph{International conference on machine learning}, 2952--2960.
  PMLR.

\bibitem[{Mnih et~al.(2016)Mnih, Badia, Mirza, Graves, Lillicrap, Harley,
  Silver, and Kavukcuoglu}]{mnih2016asynchronous}
Mnih, V.; Badia, A.~P.; Mirza, M.; Graves, A.; Lillicrap, T.; Harley, T.;
  Silver, D.; and Kavukcuoglu, K. 2016.
\newblock Asynchronous methods for deep reinforcement learning.
\newblock In \emph{International conference on machine learning}, 1928--1937.
  PMLR.

\bibitem[{Mnih et~al.(2015)Mnih, Kavukcuoglu, Silver, Rusu, Veness, Bellemare,
  Graves, Riedmiller, Fidjeland, Ostrovski et~al.}]{mnih2015human}
Mnih, V.; Kavukcuoglu, K.; Silver, D.; Rusu, A.~A.; Veness, J.; Bellemare,
  M.~G.; Graves, A.; Riedmiller, M.; Fidjeland, A.~K.; Ostrovski, G.; et~al.
  2015.
\newblock Human-level control through deep reinforcement learning.
\newblock \emph{nature}, 518(7540): 529--533.

\bibitem[{Parker-Holder, Nguyen, and Roberts(2020)}]{parker2020provably}
Parker-Holder, J.; Nguyen, V.; and Roberts, S.~J. 2020.
\newblock Provably efficient online hyperparameter optimization with
  population-based bandits.
\newblock \emph{Advances in Neural Information Processing Systems}, 33.

\bibitem[{Paul, Kurin, and Whiteson(2019)}]{NEURIPS2019_743c41a9}
Paul, S.; Kurin, V.; and Whiteson, S. 2019.
\newblock Fast Efficient Hyperparameter Tuning for Policy Gradient Methods.
\newblock In Wallach, H.; Larochelle, H.; Beygelzimer, A.; d\textquotesingle
  Alch\'{e}-Buc, F.; Fox, E.; and Garnett, R., eds., \emph{Advances in Neural
  Information Processing Systems}, volume~32. Curran Associates, Inc.

\bibitem[{Peters and Schaal(2006)}]{peters2006policy}
Peters, J.; and Schaal, S. 2006.
\newblock Policy gradient methods for robotics.
\newblock In \emph{2006 IEEE/RSJ International Conference on Intelligent Robots
  and Systems}, 2219--2225. IEEE.

\bibitem[{Pritzel et~al.(2017)Pritzel, Uria, Srinivasan, Badia, Vinyals,
  Hassabis, Wierstra, and Blundell}]{pritzel2017neural}
Pritzel, A.; Uria, B.; Srinivasan, S.; Badia, A.~P.; Vinyals, O.; Hassabis, D.;
  Wierstra, D.; and Blundell, C. 2017.
\newblock Neural episodic control.
\newblock In \emph{Proceedings of the 34th International Conference on Machine
  Learning-Volume 70}, 2827--2836. JMLR. org.

\bibitem[{Rana et~al.(2017)Rana, Li, Gupta, Nguyen, and
  Venkatesh}]{rana2017high}
Rana, S.; Li, C.; Gupta, S.; Nguyen, V.; and Venkatesh, S. 2017.
\newblock High dimensional Bayesian optimization with elastic Gaussian process.
\newblock In \emph{International conference on machine learning}, 2883--2891.
  PMLR.

\bibitem[{Schulman et~al.(2017)Schulman, Wolski, Dhariwal, Radford, and
  Klimov}]{schulman2017proximal}
Schulman, J.; Wolski, F.; Dhariwal, P.; Radford, A.; and Klimov, O. 2017.
\newblock Proximal policy optimization algorithms.
\newblock \emph{arXiv preprint arXiv:1707.06347}.

\bibitem[{Silver et~al.(2017)Silver, Schrittwieser, Simonyan, Antonoglou,
  Huang, Guez, Hubert, Baker, Lai, Bolton et~al.}]{silver2017mastering}
Silver, D.; Schrittwieser, J.; Simonyan, K.; Antonoglou, I.; Huang, A.; Guez,
  A.; Hubert, T.; Baker, L.; Lai, M.; Bolton, A.; et~al. 2017.
\newblock Mastering the game of go without human knowledge.
\newblock \emph{nature}, 550(7676): 354--359.

\bibitem[{Snoek, Larochelle, and Adams(2012)}]{snoek2012practical}
Snoek, J.; Larochelle, H.; and Adams, R.~P. 2012.
\newblock Practical Bayesian optimization of machine learning algorithms.
\newblock In \emph{Proceedings of the 25th International Conference on Neural
  Information Processing Systems-Volume 2}, 2951--2959.

\bibitem[{Sutton(1992)}]{sutton1992adapting}
Sutton, R.~S. 1992.
\newblock Adapting bias by gradient descent: An incremental version of
  delta-bar-delta.
\newblock In \emph{AAAI}, 171--176. San Jose, CA.

\bibitem[{Tang and Choromanski(2020)}]{tang2020online}
Tang, Y.; and Choromanski, K. 2020.
\newblock Online hyper-parameter tuning in off-policy learning via evolutionary
  strategies.
\newblock \emph{arXiv preprint arXiv:2006.07554}.

\bibitem[{Tulving(2002)}]{tulving2002episodic}
Tulving, E. 2002.
\newblock Episodic memory: From mind to brain.
\newblock \emph{Annual review of psychology}, 53(1): 1--25.

\bibitem[{Wu et~al.(2017)Wu, Mansimov, Liao, Grosse, and Ba}]{wu2017scalable}
Wu, Y.; Mansimov, E.; Liao, S.; Grosse, R.; and Ba, J. 2017.
\newblock Scalable trust-region method for deep reinforcement learning using
  Kronecker-factored approximation.
\newblock In \emph{Proceedings of the 31st International Conference on Neural
  Information Processing Systems}, 5285--5294.

\bibitem[{Xu, van Hasselt, and Silver(2018)}]{xu2018meta}
Xu, Z.; van Hasselt, H.~P.; and Silver, D. 2018.
\newblock Meta-Gradient Reinforcement Learning.
\newblock \emph{Advances in Neural Information Processing Systems}, 31:
  2396--2407.

\bibitem[{Young et~al.(2015)Young, Rose, Karnowski, Lim, and
  Patton}]{young2015optimizing}
Young, S.~R.; Rose, D.~C.; Karnowski, T.~P.; Lim, S.-H.; and Patton, R.~M.
  2015.
\newblock Optimizing deep learning hyper-parameters through an evolutionary
  algorithm.
\newblock In \emph{Proceedings of the Workshop on Machine Learning in
  High-Performance Computing Environments}, 1--5.

\bibitem[{Zhang et~al.(2021)Zhang, Rajan, Pineda, Lambert, Biedenkapp, Chua,
  Hutter, and Calandra}]{zhang2021importance}
Zhang, B.; Rajan, R.; Pineda, L.; Lambert, N.; Biedenkapp, A.; Chua, K.;
  Hutter, F.; and Calandra, R. 2021.
\newblock On the importance of hyperparameter optimization for model-based
  reinforcement learning.
\newblock In \emph{International Conference on Artificial Intelligence and
  Statistics}, 4015--4023. PMLR.

\bibitem[{Ziegler et~al.(2019)Ziegler, Stiennon, Wu, Brown, Radford, Amodei,
  Christiano, and Irving}]{ziegler2019fine}
Ziegler, D.~M.; Stiennon, N.; Wu, J.; Brown, T.~B.; Radford, A.; Amodei, D.;
  Christiano, P.; and Irving, G. 2019.
\newblock Fine-tuning language models from human preferences.
\newblock \emph{arXiv preprint arXiv:1909.08593}.

\end{thebibliography}

\cleardoublepage{}

\section*{Appendix}

\subsection{A. Details of methodology }

\subsubsection{A.1 Hyper-reward design\label{subsec:Hyper-reward-design}}

Each step of the Hyper-RL requires a hyper-reward. The hyper-reward
reflects how well the hyper-agent is performing to help the RL agent
in the main RL's environment. The tricky part is the RL agent's performance
is not always measured at every Hyper-RL step (policy update step). 

We define $\mathtt{U}$ the interval of Hyper-RL steps (the update
phase) between 2 performance measurement of the RL agent. At the end
of each update phase, after taking hyper-action and update models,
the performance is evaluated in the environment phase, resulting in
the roll-out return $G$, which will be used as hyper-reward for the
final step of the update phase. For in-between steps in the update
phase, there is no direct way to know the intermediate outcome of
each hyper-action, hence a hyper-reward $0$ is assigned, making Hyper-RL
generally a sparse problem.

One may think of assigning in-between steps the same hyper-return,
which is collected from the previous/next environment phase to avoid
sparse hyper-reward. However, it does not make sense to use past/future
outcomes to assign reward for current steps. Hence, we choose the
sparse reward scheme and that is also one motivation for using episodic
memory. 

Since EPGT reuses PG return, it does not require extra computation
for hyper-reward, unlike HOOF (recomputing returns with important
sampling) or population-based methods (collecting return in parallel).

\subsubsection{A.2 Hyper-action quantization\label{subsec:Hyper-action-quantization}}

For most types of hyperparameters, we use uniform quantization within
a range around the default value to derive the hyper-action. For example,
when $\mathtt{B}=3$, GAE's $\lambda$ will have the possible values
$\left\{ 0.95,0.975,0.99\right\} $ and PPO's clip $\epsilon\in$
$\left\{ 0.1,0.2,0.3\right\} $

For learning rate, uniform quantization for a certain range will not
necessarily include the default learning rate, which can be a good
candidate for the hyperparameter. Hence, we use a different quantization
formula as follows,

\begin{equation}
\left\{ \frac{\alpha^{*}}{\left(\mathtt{B}-1\right)/2},...,\frac{\alpha^{*}}{2},\alpha^{*},\alpha^{*}\times2,...,\alpha^{*}\times\left(B-1\right)/2\right\} \label{eq:hpa}
\end{equation}
where $\alpha^{*}$ is the default learning rate (for A2C, e.g. $\alpha^{*}=7\times10^{-4}$).
This is convenient for automatically generating the bins for hyper-action
space given $\mathtt{B}$, which ensures that there exists one action
that correspond to the default hyperparameter. 

One limitation of discretizing hyper-actions is the exponential growth
of the number of hyper-action w.r.t the size of the optimized hyperparameter
set. For example, if there are $\left\Vert \psi\right\Vert $ hyperparameters
we want to optimize on-the-fly and each is quantized into $\mathtt{B_{i}}$
discrete values, the number of hyper-actions is 

\[
\mathtt{\left\Vert A\right\Vert =\mathtt{\prod_{i}^{\left\Vert \psi\right\Vert }B_{i}}}
\]

In this paper, our experiments scale up to $\left\Vert \mathtt{A}\right\Vert =3\times3\times4\times4=144$.
Beyond this limit may require a different way to model the hyper-action
space (e.g. RL methods for continuous action space). 

\subsubsection{A.3 EPGT's networks\label{subsec:EPO's-networks}}

EPGT has trainable parameters, which are $C_{m}^{n}$, $\phi$ and
$\omega$. Here, $C_{m}^{n}$ is just a parametric 2d tensor, $\phi$
and $\omega$ are neural networks, implemented as follows:
\begin{itemize}
\item Encoder network $\phi$: 2-layer feed-forward neural network with
$\tanh$ activation with layer size: $d\rightarrow d/4\rightarrow h\times2$.
The output will be used as the mean and the standard deviation of
the normal distribution in VAE.
\item Decoder network $\omega$: 2-layer feed-forward neural network with
$\sigmoid$ activation with layer size: $h\rightarrow d/4\rightarrow d$
\end{itemize}
The number of parameters will increase as the PG networks grow. For
PPO as the most complicated example, EPGT's number of parameters is
5M.

\begin{figure*}
\begin{centering}
\includegraphics[width=1\textwidth]{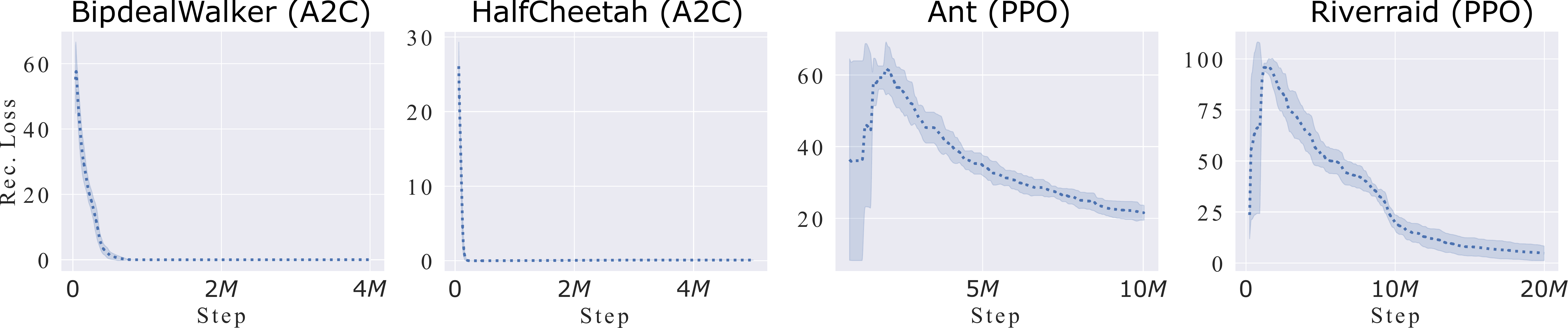}
\par\end{centering}
\caption{Reconstruction loss training VAE in representative environments and
PG methods. The curves are mean with std. over 5 runs. \label{fig:Reconstruction-loss-training}}
\end{figure*}

\begin{figure*}
\begin{centering}
\includegraphics[width=1\textwidth]{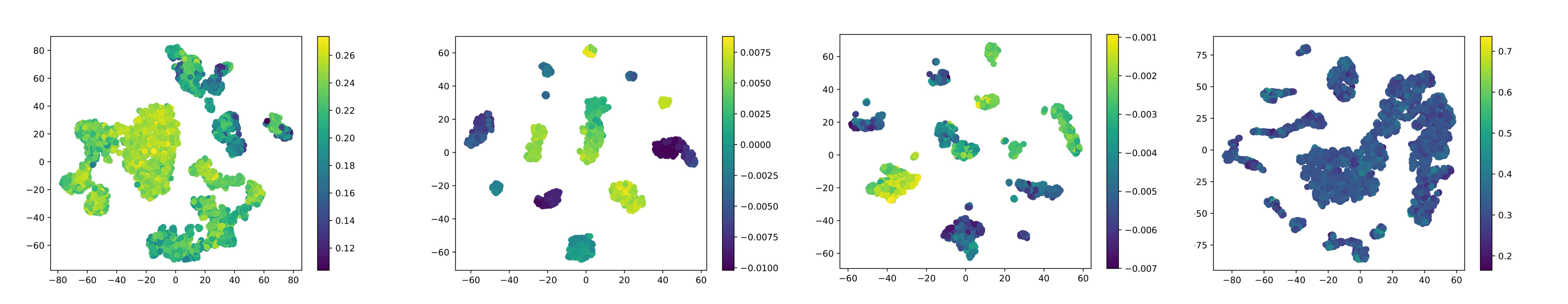}
\par\end{centering}
\caption{Hyper-state representations stored in memory for BipedalWalker, HalfCheetah,
Ant, Riverraid tasks (from left to right) at 1000 update steps. \label{fig:Reconstruction-loss-training-1}}
\end{figure*}

To train the networks, we sample hyper-states in the buffer $\mathtt{D}$
and minimize $\mathcal{L}_{rec}=\left\Vert \omega\left(\phi\left(\mathtt{s}^{\psi}\right)\right)-\mathtt{s^{\psi}}\right\Vert _{2}^{2}$
using gradient decent and batch size of 8. Here, we stop the gradient
at the target $\mathtt{s}^{\psi}$ and only let the gradient backpropagated
via $\omega\left(\phi\left(\mathtt{s^{\psi}}\right)\right)$ to learn
$\omega$,$\phi$ and $\left\{ C_{m}^{n}\right\} _{n=0,m=1}^{N_{order},M}$.
In our experiments, to save computing cost, we do not sample every
learning step. Instead, every 10 policy update steps, we sample data
and perform a gradient decent step to minimize $\mathcal{L}_{rec}$
once. Training, interestingly, is usually sample-efficient, especially
when $d$ is not big and PG networks are simple. Fig. \ref{fig:Reconstruction-loss-training}
showcases the reconstruction loss $\mathcal{L}_{rec}$ in several
environments using PG A2C and PPO. For PG A2C, convergence is quickly
achieved since A2C's network for Bipedal Walker and HalfCheetah is
simple. For PPO, the task is more challenging and also the networks
include CNN (in case of Atari game Riverraid). Hence, the loss is
minimized slower, yet still showing good convergence. We note that
the model learn meaning representations and the mapping is not degenerated.
According to Fig. \ref{fig:(a)-Learning-state}, our learned mapping
distributes states to clusters, showing clear discrimination between
similar and dissimilar states (see more in Fig. \ref{fig:Reconstruction-loss-training-1}). 

\subsection{B. Details of experiments\label{subsec:Details-of-experiments}}

\subsubsection{B.1 A summary of tasks and PG methods}

We use environments from Open AI gyms \footnote{\url{https://gym.openai.com/envs/#classic_control}},
which are public and using The MIT License. Mujoco environments use
Mujoco software\footnote{\url{https://www.roboti.us/license.html}}
(our license is academic lab). Table \ref{tab:Tasks-used-in} lists
all the environments. 

All PG methods (A2C, ACKTR, PPO) use available public code. They are
Pytorch reimplementation of OpenAI's stable baselines\footnote{\url{https://github.com/ikostrikov/pytorch-a2c-ppo-acktr-gail}},
which can reproduce the original performance relatively well. The
source code for baselines HOOF and PB2 is adopted from the authors'
public code \footnote{\url{https://github.com/supratikp/HOOF}} and
\footnote{\url{https://github.com/jparkerholder/PB2}}, respectively.
Table \ref{tab:PGs-used-in} summarizes the PG methods and their networks. 

\begin{table}
\begin{centering}
{\footnotesize{}}%
\begin{tabular}{ccc}
\hline 
\multirow{2}{*}{\textbf{\footnotesize{}Tasks}} & \textbf{\footnotesize{}Continuous } & \textbf{\footnotesize{}Gym }\tabularnewline
 & \textbf{\footnotesize{}action} & \textbf{\footnotesize{}category}\tabularnewline
\hline 
{\footnotesize{}Mountain Car-v0} & {\footnotesize{}$\mathit{\mathtt{X}}$} & {\footnotesize{}Classical }\tabularnewline
{\footnotesize{}Mountain Car Continuous-v0} & {\footnotesize{}$\checked$} & {\footnotesize{}control}\tabularnewline
\hline 
{\footnotesize{}Bipedal Walker-v3} & \multirow{2}{*}{{\footnotesize{}$\checked$}} & \multirow{2}{*}{{\footnotesize{}Box2d}}\tabularnewline
{\footnotesize{}Lunar Lander Continuous-v2} &  & \tabularnewline
\hline 
{\footnotesize{}MuJoCo tasks (v2): HalfCheetah} & \multirow{3}{*}{{\footnotesize{}$\checked$}} & \multirow{3}{*}{{\footnotesize{}MuJoCo}}\tabularnewline
{\footnotesize{}Walker2d, Hopper, Ant} &  & \tabularnewline
{\footnotesize{}Swimmer, Humanoid,} &  & \tabularnewline
\hline 
{\footnotesize{}Atari games (NoFramskip-v4):} & \multirow{5}{*}{{\footnotesize{}$\mathit{\mathtt{X}}$}} & \multirow{5}{*}{{\footnotesize{}Atari}}\tabularnewline
{\footnotesize{}Beamrider, Breakout, Pong} &  & \tabularnewline
{\footnotesize{}Qbert, Seaquest, SpaceInvaders} &  & \tabularnewline
{\footnotesize{}Alien, MsPacman, Demonattack} &  & \tabularnewline
{\footnotesize{}Riverraid} &  & \tabularnewline
\hline 
\end{tabular}{\footnotesize\par}
\par\end{centering}
\caption{Tasks used in the paper. \label{tab:Tasks-used-in}}
\end{table}
\begin{table}
\begin{centering}
{\footnotesize{}}%
\begin{tabular}{cc}
\hline 
\textbf{\footnotesize{}PGs} & \textbf{\footnotesize{}Policy/Value networks}\tabularnewline
\hline 
\multirow{5}{*}{{\footnotesize{}A2C/ACKTR/PPO}} & \emph{\footnotesize{}Vector input:}{\footnotesize{} 2-layer feedforward }\tabularnewline
 & {\footnotesize{}net (tanh, h=32)}\tabularnewline
 & \emph{\footnotesize{}Image input:}{\footnotesize{} 3-layer ReLU CNN
with}\tabularnewline
 & {\footnotesize{}kernels $\left\{ 32/8/4,64/4/2,32/3/1\right\} $+2-layer}\tabularnewline
 & {\footnotesize{}feedforward net (ReLU, h=512)}\tabularnewline
\hline 
\end{tabular}{\footnotesize\par}
\par\end{centering}
\caption{PG methods used in the paper. \label{tab:PGs-used-in}}
\end{table}

\subsubsection{B.2 Episodic memory configuration}

We implement the memory using kd-tree structure, so neighbor lookup
for memory $\mathtt{read}$ and $\mathtt{update}$ is fast. To further
reduce computation complexity, we only update the memory after every
10 policy update steps. The memory itself has many hyperparameters.
However, to keep our solution efficient, we keep most of the hyperparameters
unchanged across experiments and do not tune them. In particular,
the list of untuned hyperparameters is:
\begin{itemize}
\item Hyperparameters of the VAE (see A.3)
\item State embedding size $h=32$
\item The writing rate $\beta=0.5$
\item Similarity kernel $Sim\left(i,k\right)=\frac{1}{\left\Vert \phi\left(\mathtt{s}_{k}^{\psi}\right)-\phi\left(\mathtt{s}_{i}^{\psi}\right)\right\Vert +\epsilon}$,
$\epsilon=0.001$ following \cite{pritzel2017neural}
\item $\epsilon$ in the $\epsilon$-greedy is linearly reduced from 1.0
to 0 during the training. At the final step of training, $\epsilon=0$.
\item The memory size $N_{mem}$ is always half of the number of policy
update steps for each task. This is determined by our educated guess
that after half of training time, the initial learning experiences
stored in the memory is not relevant anymore and needed to be replace
with newer observations. The number of policy update steps can be
always computed given the allowed number of environment steps. For
example, if we allow $n_{e}$ environment steps for training, then
the number of policy updates is $\frac{n_{e}\times\mathtt{U}}{T}$
, and the memory size will be $\frac{n_{e}\times\mathtt{U}}{20T}$
(since we update every 10 steps). Here, we assume each memory update
will add a new tuple to the memory since the chance of exact key match
is very small. For all of our tasks, the maximum $N_{mem}$ is up
to 10,000. 
\item To hasten the training with EPGT, we utilize uniform writing \cite{le2019learning}
to reduce the compute complexity of memory writing operator. In particular,
we write observations to the episodic memory every 10 update steps.
\end{itemize}
For other hyperparameters of EPGT, we tune or verify them in Mountain
Car or one Atari game and apply the found hyperparameter values to
all other tasks. These hyperparameters are:
\begin{itemize}
\item VAE or Random projection, verified in Mountain Car
\item Order of representation $N_{order}$, tuned in MsPacman
\item State size $d$, tuned in MsPacman
\item Neighbor size $K$, tuned in Demon Attack and Alien
\end{itemize}
The details of selecting values for these hyperparameters can be found
in ablation studies in this paper. We note that after verifying reasonable
values for these EPGT hyperparameters in some environments, we keep
using them across all experiments, and thus do not requires additional
tunning per task.

\subsubsection{B.3 Training description\label{subsec:Training-description}}

All the environments are adopted from Open AI's Gym (MIT license).

\paragraph{Mountain Car Continuous and Bipedal Walker with PG A2C}

We use the A2C with default hyperparameters from Open AI\footnote{\url{https://stable-baselines.readthedocs.io/en/master/modules/a2c.html}}.
In these two tasks, we optimize the learning rate $\alpha$ from the
range defined by Eq. \ref{eq:hpa} using $\mathtt{B}=15$ and $\mathtt{B=5}$,
respectively. Here, $\mathtt{U=10}$. In this task, we implement 2
additional baselines solving the Hyper-RL: DQN and Random (RND) agent.
Both DQN and RND uses the same hyper-state (trained with VAE), action
and reward representation as EPGT. We note that these baselines are
used to optimize hyperparameters of the PG algorithm, not for solving
the main RL.

For RND, we uniformly sample the hyper-action at each policy update
step, which is equivalent to a random hyper-agent. For DQN\footnote{Implementation from public source code \url{https://github.com/higgsfield/RL-Adventure}},
we also use $\epsilon$-greedy is linearly reduced from 1.0 to 0 during
the training and the value and target networks are 3-layer ReLU feedforward
net with $64$ hidden units. DQN is trained with a replay buffer size
of 1 million and batch size of 32. The DQN's networks are updated
at every policy update step and the target network is synced with
the value network every $T_{q}$ policy update steps. We tune $T_{q}$
with difference values (5, 50, 500) for each task and report the best
performance.

To measure the computing efficiency, we compare the running speed
and memory usage of EPGT, DQN (Hyper-RL) and the original A2C in Table
\ref{tab:Computing-cost-of}. On our machines using 1 GPU Tesla V100-SXM2,
in terms of speed, EPGT runs slightly slower than A2C without any
scheduler, yet much faster than DQN. When the problem gets complicated
as in BW, EPGT is twice faster than DQN. In terms of memory, all models
consume similar amount of memory. EPGT is slightly less RAM-consuming
than DQN since the episodic memory size is smaller than the number
of parameters of the DQN's networks. 

\begin{table}
\begin{centering}
\begin{tabular}{cccc}
\hline 
\multirow{2}{*}{Model} & \multicolumn{2}{c}{Speed (env. steps/s)} & \multicolumn{1}{c}{Mem (Gb)}\tabularnewline
\cline{2-4} 
 & MCC & BW & MCC and BW\tabularnewline
\hline 
Default (A2C) & 780  & 78 & 1.49\tabularnewline
DQN (Hyper-RL) & 590 & 35 & 1.59\tabularnewline
\hline 
EPGT & 720 & 71 & 1.58\tabularnewline
\hline 
\end{tabular}
\par\end{centering}
\caption{Computing cost of different methods in MountainCarContinuous (MCC)
and BipedalWalker (BW). \label{tab:Computing-cost-of}}
\end{table}

\paragraph{4 Mujoco tasks with PG A2C}

In this task, we follow closely the training setting in \cite{NEURIPS2019_743c41a9}
using the benchmark of 4 Mujoco tasks and optimizing learning rate
$\alpha$. The PG method is A2C, trained on 5 million environment
steps. The configuration of A2C is similar to that of \cite{NEURIPS2019_743c41a9}
($T=5$$,$RMSProb optimizer with initial learning rate of $7\times10^{-4}$,
value loss coefficient of 0.5, entropy loss coefficient of 0.01, GAE
$\lambda=0.95$, $\gamma=0.99$, etc. ). Here, $\mathtt{U=1}$. The
learning rate $\alpha$ is optimized in the range defined by Eq. \ref{eq:hpa}
using $\mathtt{B}=7$.

\begin{figure*}
\begin{centering}
\includegraphics[width=1\textwidth]{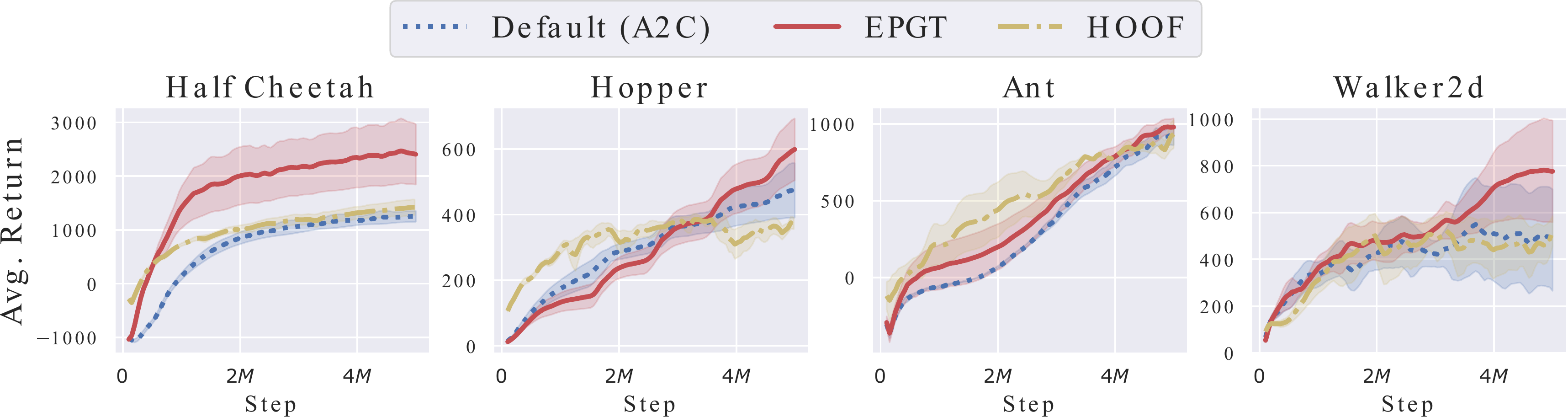}
\par\end{centering}
\caption{4 Mujoco tasks: Average return over environmental steps across 10
training seeds using PG A2C.\label{fig:Average-return-over}}
\end{figure*}

Fig. \ref{fig:Average-return-over} compares the learning curves of
EPGT against the baseline A2C with default hyperparameters. EPGT improves
A2C performance by a huge margin in HalfCheetah and Walker2d. The
other two tasks show smaller improvement. We report EPGT's numbers
in Table \ref{tab:EPO-vs-sequential} by using the best checkpoint
to measure average return over 100 episodes for each run, then take
average over 10 runs. 

\paragraph{4 mixed tasks with PG PPO}

We use similar setting introduced in \cite{parker2020provably} using
4 tasks: BipedalWalker, LunarLanderContinous, Hopper and InvertedDoublePendulum,
each is trained using 1 million environment steps. Here, 4 hyperparameters
are optimized: learning rate $\alpha$, batch size $b$, GAE $\lambda$
and PPO clip $\epsilon$. The PG algorithm is PPO with configuration:
$T=1600$, number of gradient updates=10, num workers=4, Adam optimizer
with initial learning rate of $3\times10^{-4}$ and $\gamma=0.99$
. Here, $\mathtt{T_{u}}=10\times4\times1600/b$ where $b$ is the
batch size. The range of optimized hyperparameters: $b$ $\left\{ 128,256,512\right\} $,
$\lambda$ $\left\{ 0.9,0.95,0.975,0.99\right\} $, $\epsilon$ $\left\{ 0.1,0.2,0.3,,0.5\right\} $,
$\alpha$ using Eq. \ref{eq:hpa} with $\mathtt{B}=3$.

\begin{figure*}
\begin{centering}
\includegraphics[width=1\textwidth]{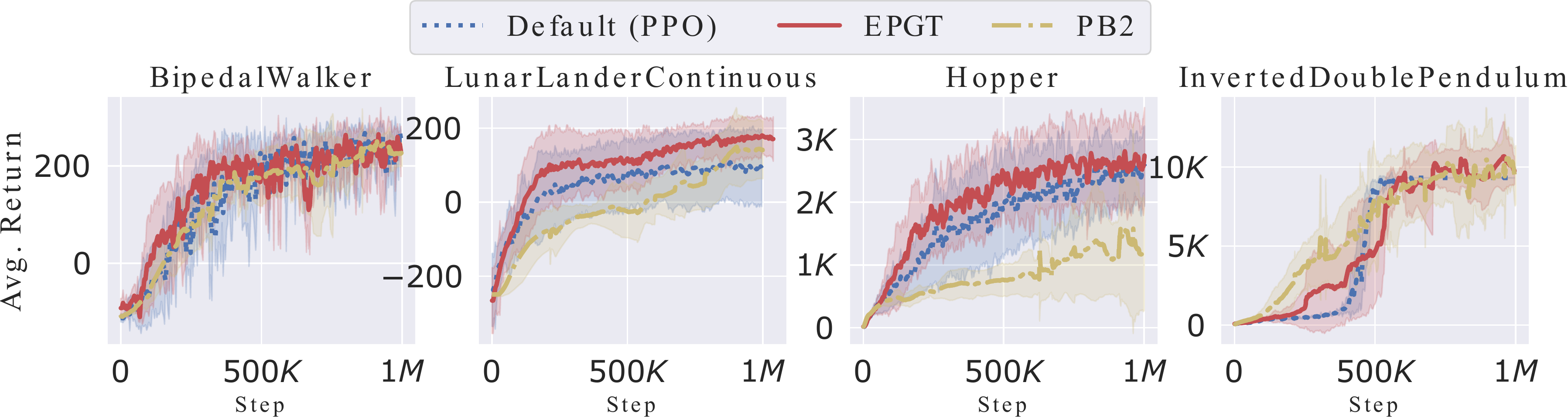}
\par\end{centering}
\caption{4 tasks: Average return over environment steps across 10 training
seeds using PG PPO.\label{fig:Average-return-over-1}}
\end{figure*}

Fig. \ref{fig:Average-return-over-1} compares the learning curves
of EPGT against the baseline PPO with default hyperparameters. EPGT
shows clear improvement in LunarLanderContinous and Hopper. We report
EPGT's numbers in Table \ref{tab:EPO-vs-parallel} by using the best
checkpoint to measure average return over 100 episodes for each run,
then take median over 10 runs. 

\begin{figure*}
\begin{centering}
\includegraphics[width=1\textwidth]{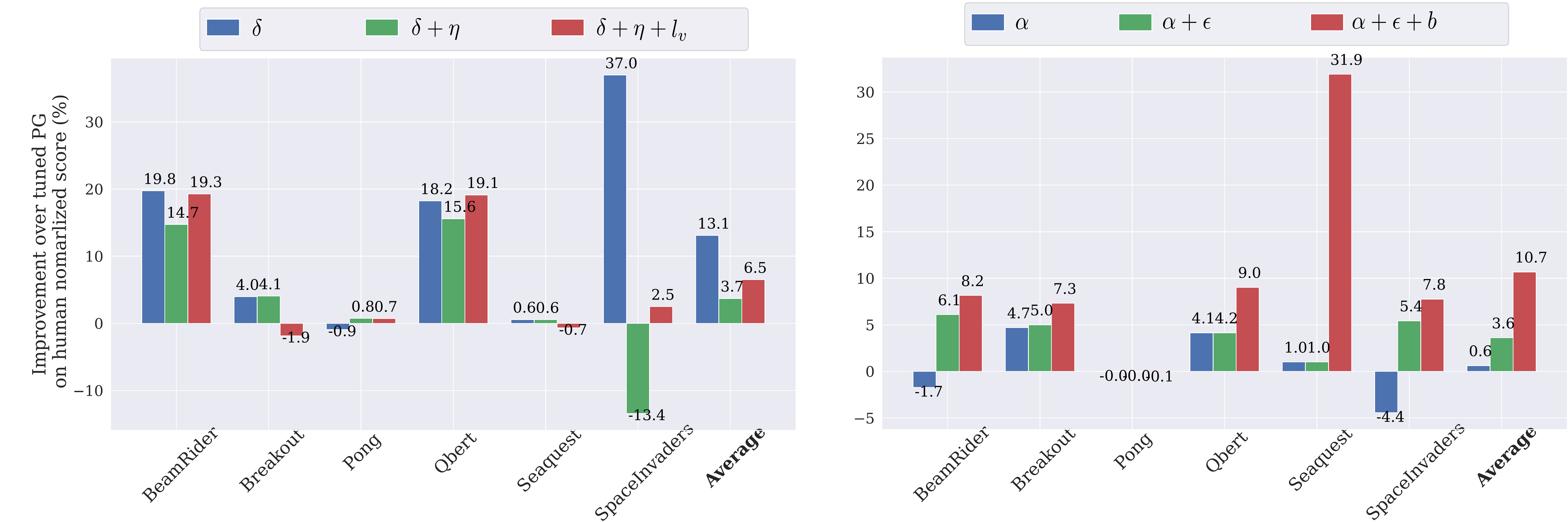}
\par\end{centering}
\caption{Atari games: Average return over environment steps across 5 training
seeds using PG ACKTR (left) amd PPO (right).\label{fig:Atari-games:-Average-1-1}}
\end{figure*}

\paragraph{Atari tasks with ACKTR}

We adopt ACKTR as PG method with default hyperparameters \cite{wu2017scalable}:
number of workers=40, initial $\eta=0.25$, GAE $\lambda=0.95$, $\gamma=0.99$,
$T=20$ and $\mathtt{U=10}$. The full set of tuned hyperparameters
is trust region radius $\delta$ $\left\{ 0.001,0.002,0.003\right\} $,
the value loss coefficient $l_{v}$$\left\{ 0.25,0.5,1.0\right\} $
and step size $\eta$ using Eq. \ref{eq:hpa} with $\mathtt{B}=3$.

\begin{figure*}
\begin{centering}
\includegraphics[width=1\textwidth]{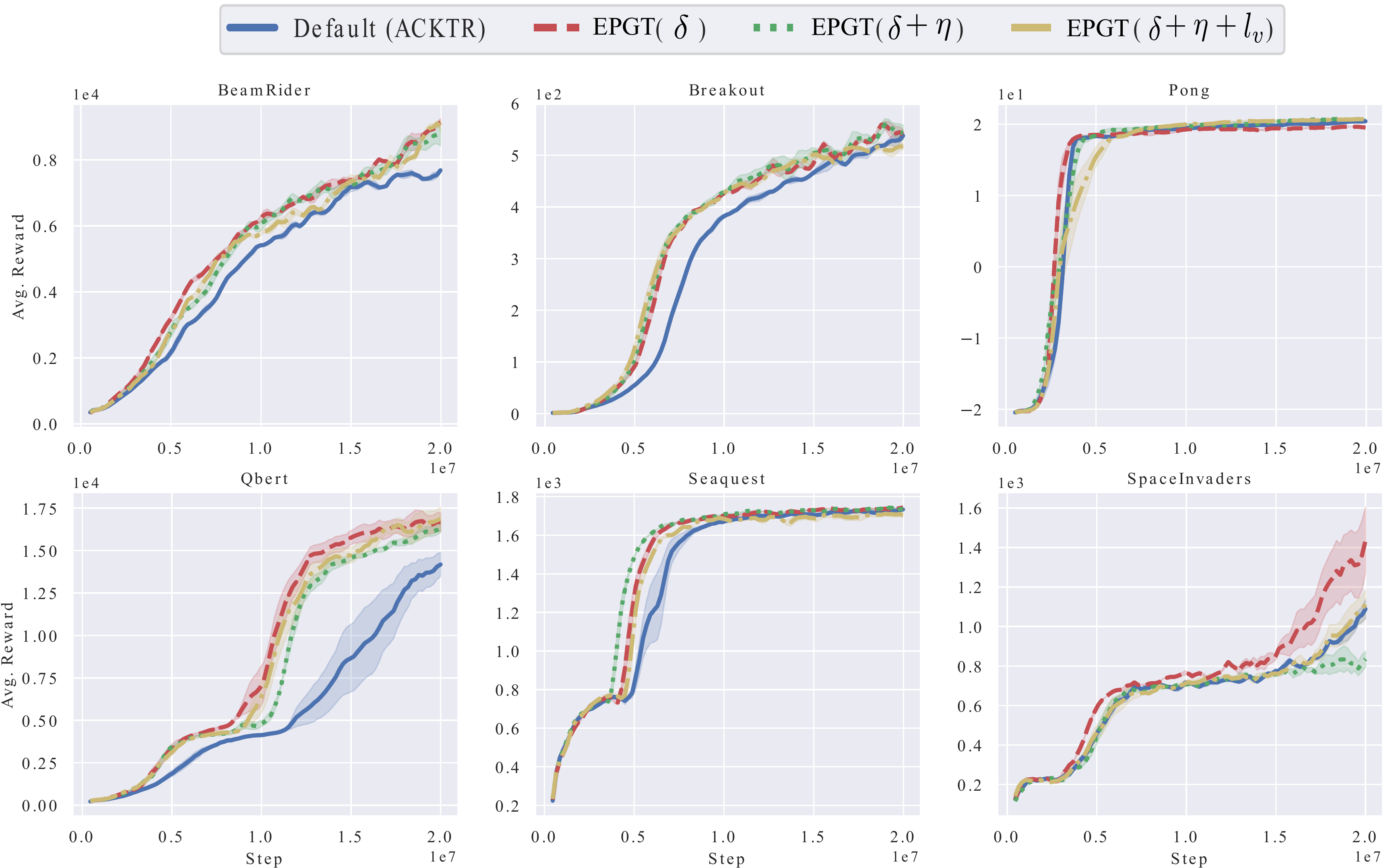}
\par\end{centering}
\caption{Atari games: Average return over environment steps across 5 training
seeds using PG ACKTR.\label{fig:Atari-games:-Average-1}}
\end{figure*}

Fig. \ref{fig:Atari-games:-Average-1} compares the learning curves
of EPGT with different optimized hyperparameter set against the best
ACKTR. The best ACKTR is found by grid-search ($\delta$ $\left\{ 0.001,0.003\right\} $,
$l_{v}$$\left\{ 0.5,1.0\right\} $, $\eta\left\{ 0.2,0.07\right\} $)
on Breakout. EPGT shows clear improvement in BeamRider, Qbert and
SpaceInvaders. 

\paragraph{Atari tasks with PPO}

We adopt PPO as PG method with default hyperparameters \cite{schulman2017proximal}:
$T=2048$, number of gradient updates=10, num workers=1, Adam optimizer
with initial learning rate of $3\times10^{-4}$ and $\gamma=0.99$
. Here, $\mathtt{U}=10\times4\times2048/b$ where $b$ is the batch
size. The range of optimized hyperparameters: $b$ $\left\{ 128,256,512\right\} $,
$\lambda$ $\left\{ 0.9,0.95,0.975,0.99\right\} $, $\epsilon$ $\left\{ 0.1,0.2,0.3,0.5\right\} $,
$\alpha$ using Eq. \ref{eq:hpa} with $\mathtt{B}=3$.

\begin{figure*}
\begin{centering}
\includegraphics[width=1\textwidth]{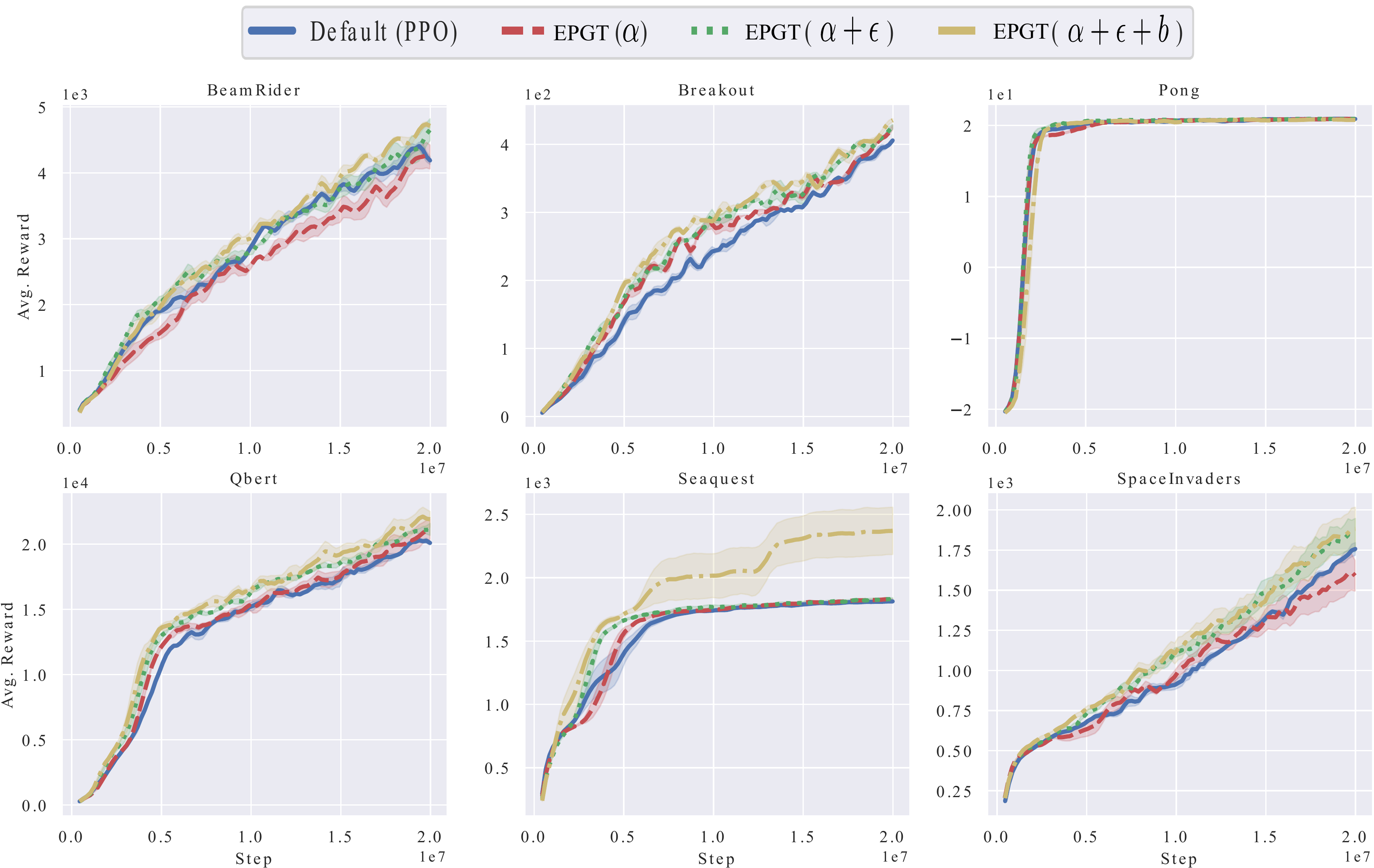}
\par\end{centering}
\caption{Atari games: Average return over environment steps across 5 training
seeds using PG PPO.\label{fig:Atari-games:-Average}}
\end{figure*}

Fig. \ref{fig:Atari-games:-Average} compares the learning curves
of EPGT with different optimized hyperparameter set against the best
PPO. The best PPO is found by grid-search ($b$ $\left\{ 128,256\right\} $,
$\lambda$ $\left\{ 0.95,0.99\right\} $, $\epsilon$ $\left\{ 0.1,0.2\right\} $,
$\alpha\left\{ 7\times10^{-4},10^{-5}\right\} $) on Breakout. EPGT
shows clear improvement in Breakout, Qbert, Seaquest and SpaceInvaders.

We note that we only perform grid-search on smaller hyperparameter
value sets since grid-search is very expensive. EPGT only needs one
run with similar training time as the original PG methods to achieve
significantly better results than the best PG methods found after
16 runs of grid-search. 

\paragraph{6 Mujoco tasks with PPO}

We adopt PPO as PG method with default hyperparameters \cite{schulman2017proximal}:
number of workers=1, $T=2048$, $\gamma=0.99$, and Adam optimizer
with initial learning rate of $3\times10^{-4}$. Here, $\mathtt{U}=10\times1\times2048/b$.
The full set of tuned hyperparameters is $b$ $\left\{ 128,256,512\right\} $,
$\lambda$ $\left\{ 0.95,0.975,0.99\right\} $, $\epsilon$ $\left\{ 0.1,0.2,0.3\right\} $,
$\alpha$ using Eq. \ref{eq:hpa} with $\mathtt{B}=3$.

\begin{figure*}
\begin{centering}
\includegraphics[width=1\textwidth]{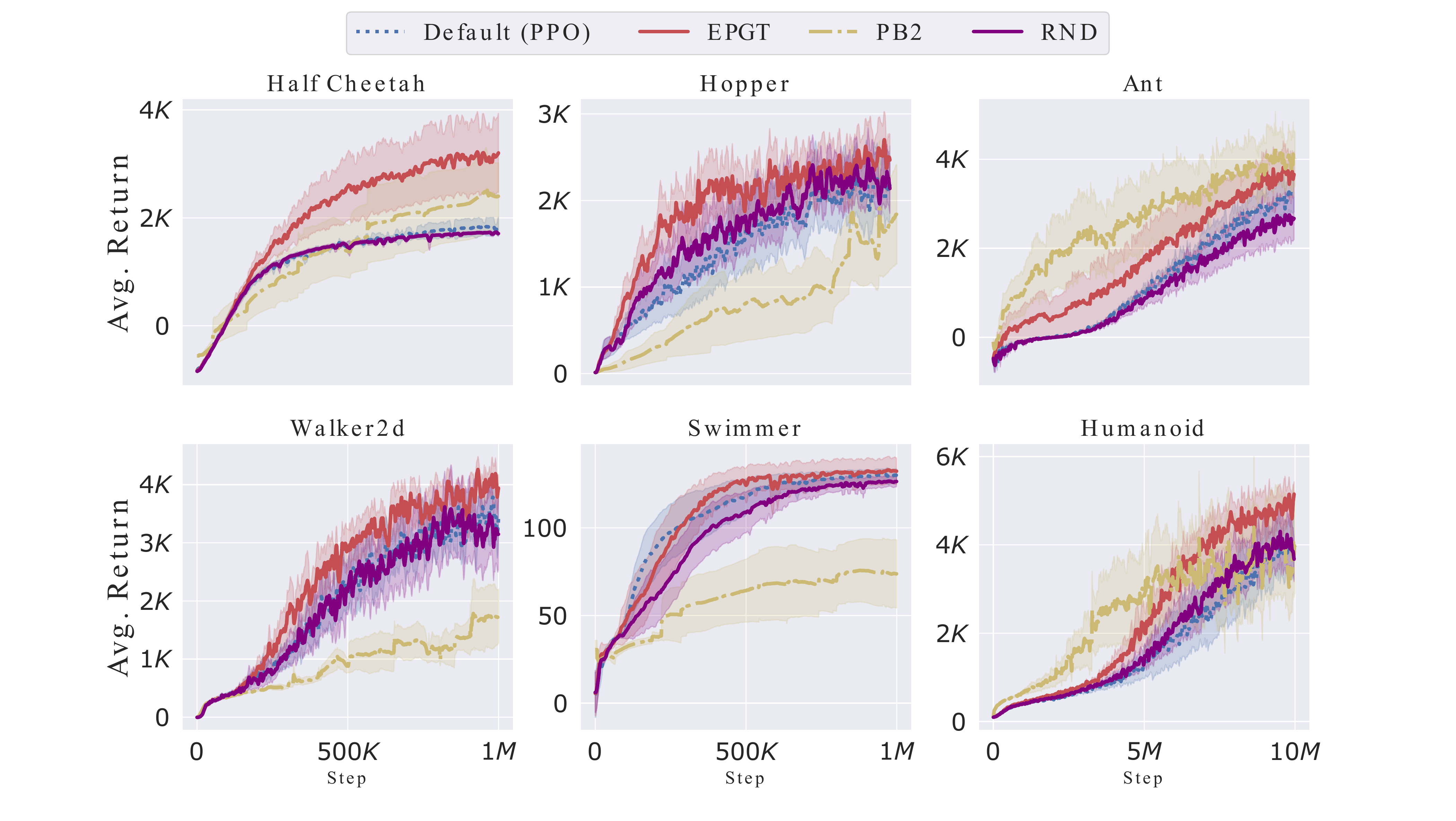}
\par\end{centering}
\caption{Mujoco: Average return over environment steps across 10 training seeds
using PG PPO.\label{fig:Mujoco-games:-Average}}
\end{figure*}

Fig. \ref{fig:Mujoco-games:-Average} compares the learning curves
of EPGT against the baseline default PPO and Random (RND) hyperparameter
selection. EPGT shows clear improvement in HalfCheetah, Hopper, Walker2d
and Humanoid. 

\subsubsection{B.4 Details of ablation studies }

\paragraph{Hyper-state size\label{par:Hyper-state-size}}

Each tensor $W_{m}^{n}$ will be projected to a vector sized $d'\times d$
where $d'$ is the first dimension of $W_{m}^{n}$ and $d$ the last
dimension of the mapping $C_{m}^{n}$. If the PG models has $M$ layers
and we maintain $N_{order}$ orders of representations, the hyper-state
vector's size, in general, is $M\times N_{order}\times d'\times d$.
We already ablate $N_{order}$ in the main manuscript, $M$ and $d'$
are fixed for each PG method, now we examine $d$.

\begin{figure*}
\begin{centering}
\includegraphics[width=1\textwidth]{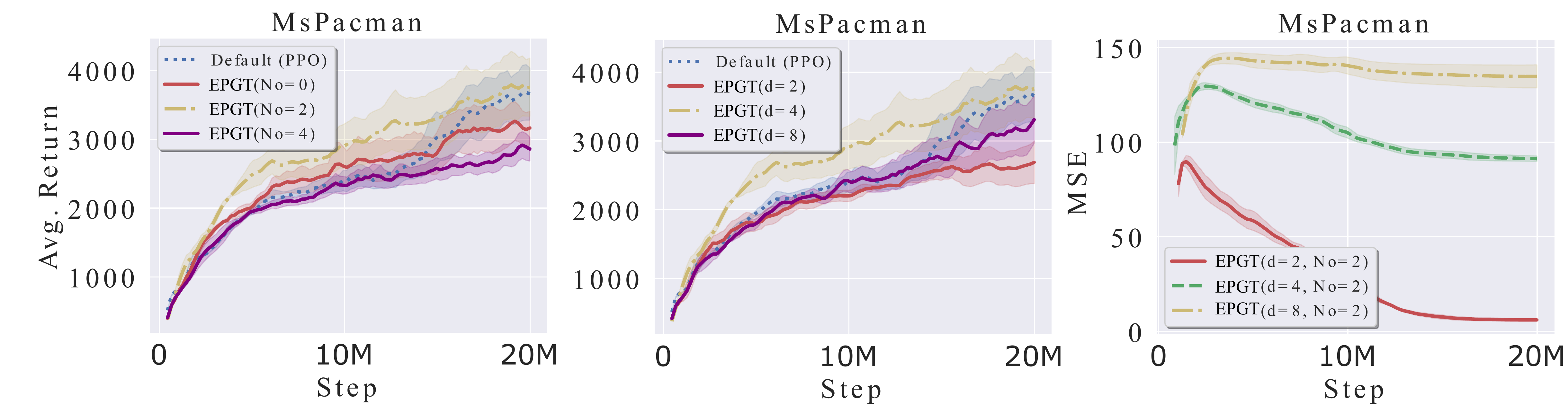}
\par\end{centering}
\caption{Ablation studies on order of representation $N_{order}$ (left) and
projected size $d$ (middle). (Right) MSE loss of VAE model for different
$d$. The curves are mean with std. over 5 runs. \label{fig:Ablation-studies-on}}
\end{figure*}

Fig. \ref{fig:Ablation-studies-on} reports the ablation results.
The common behavior is the performance improves when $N_{order}$
and $d$ increase to $N_{order}=2$ and $d=4$, which corresponds
to $8K$-dimensional hyper-state vector in this experiment. When $N_{order}=0$,
no derivative is used to represent the hyper-state, which leads to
lower performance. When either $N_{order}$ or $d$ keeps growing,
state vector size will be $16K$, which is too big for the episodic
control method to work properly. For example, VAE almost cannot learn
to minimize the reconstruction loss for $N_{order}=2$ and $d=8$
as shown in Fig. \ref{fig:Ablation-studies-on} (right).

\paragraph{Neighbor size K}

\begin{figure*}
\begin{centering}
\includegraphics[width=1\textwidth]{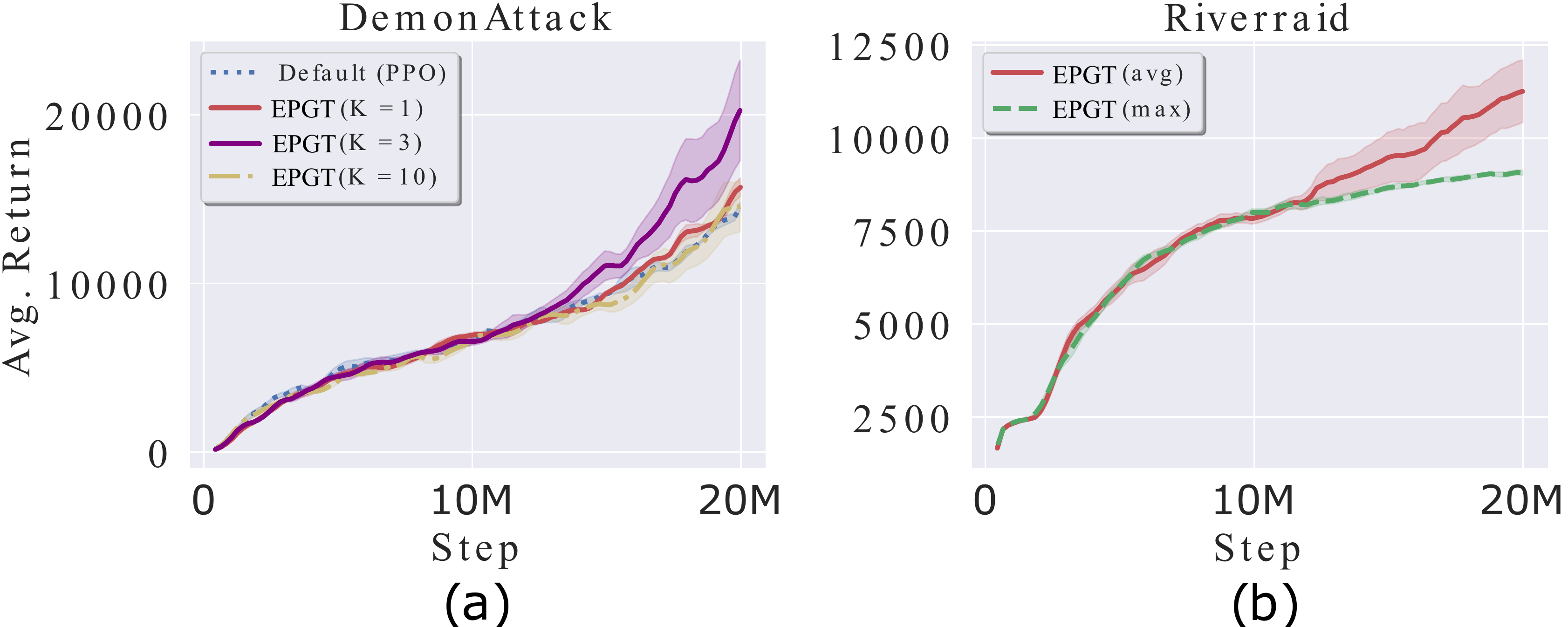}
\par\end{centering}
\caption{Ablation studies on neighbor size $K$ (a) and writing types (b).
The curves are mean with std. over 5 runs. \label{fig:Ablation-studies-on-1}}
\end{figure*}

We also test our model with different number of neighbors $K$ for
both reading and writing. We use Atari game Demon Attack as an illustration
and realize that $K=3$ is the best value (Fig. \ref{fig:Ablation-studies-on-1}
(a)). More neighbors do not help because perhaps the cluster size
of the similar-value representations is not big and thus, referring
to far neighbors is not reliable. 

\paragraph*{Average vs Max writing rule }

Traditional episodic controls \cite{blundell2016model} use max-rule
as 

\[
\mathtt{M}\left(\phi\mathtt{\left(s^{\psi}\right)},\mathtt{a}_{i}^{\psi}\right)\leftarrow\max\left(\mathtt{M}\left(\phi\mathtt{\left(s^{\psi}\right)},\mathtt{a}_{i}^{\psi}\right),\mathtt{G}_{i}^{\psi}\right)
\]
That stores the best experience the agent has so far and quickly guides
next actions toward the best experience. However, this works only
for near-deterministic environments as when $\mathtt{G}_{i}^{\psi}$
is unexpectedly high due to stochastic transition, a bad action is
misunderstood as good one. This false belief may never be updated.
Our Hyper-RL is not near-deterministic as the hyper-reward is basically
MCMC estimation of the environment returns and the state is partially
observable. Another problem of the max rule is it requires exact match
to facilitate an update, which is rare in practice. 

We address theses issue by taking weighted average and employing neighbor
writing as in Eq. \ref{eq:knnw}. As we write to multiple memory slots,
it does not require exact match and enables fast value propagation
inside the memory. Fig. \ref{fig:Ablation-studies-on-1} (b) shows
the performance increase using the average compared to max rule in
Atari Riverraid game. 

\subsection{C. Theoretical analysis of our writing rule}

In this section, we show that using our writing, the value stored
in the episodic memory is an approximation of the expected return
(our writing is generic and can be used for general RL, hence we exclude
``hyper-'' in this section). The proof is based on \cite{le2021modelbased}.
In particular, we can always find $\beta$ such that the writing converges
with probability $1$ and we also analyze the convergence as $\beta$
is constant, which is practically used in this paper.

To simplify the notation, we rewrite Eq. \ref{eq:knnw} as 

\begin{equation}
\mathtt{M}_{i}\left(n+1\right)=\mathtt{M}_{i}\left(n\right)+\text{\ensuremath{\lambda\left(n\right)}}\left(\mathtt{G}_{j}\left(n\right)-\mathtt{M}_{i}\left(n\right)\right)\label{eq:write2}
\end{equation}
where $i$ and $j$ denote the current memory slot being updated and
its neighbor that initiates the writing, respectively (it is opposite
to the indices in Eq. \ref{eq:knnw}). Here, the action is the same
for both $i$ and $j$ slots, so we exclude action to simplify the
notation. $\lambda\left(n\right)=\beta(n)\frac{Sim\left(i,j\right)}{\sum_{b=1}^{|\mathcal{N}(j)|}S_{bj}}$
where $\mathcal{N}\left(j\right)$ is the set of $K$ neighbors of
$j$ slot. $\mathtt{G}_{j}$ is the return of the state-action whose
key is the memory slot $j$, $Sim\left(j,i\right)$ the kernel function
of 2 keys and $n$ the number of memory updates. This stochastic approximation
converges when $\sum_{n=1}^{\infty}\lambda\left(n\right)=\infty$
and $\sum_{n=1}^{\infty}\lambda^{2}\left(n\right)<\infty$. 

By definition, $Sim\left(i,j\right)=\frac{1}{\left\Vert \phi\left(\mathtt{s}_{i}^{\psi}\right)-\phi\left(\mathtt{s}_{j}^{\psi}\right)\right\Vert +\epsilon}$
and $\phi\left(\mathtt{s}^{\psi}\right)\leq1$ since we use $\tanh$
activation in $\phi$. Hence, we have $\forall i,j$: $0<\frac{1}{2+\epsilon}\leq Sim\left(i,j\right)\leq\frac{1}{\epsilon}$.
Hence, let $B_{ij}\left(n\right)$ a random variable denoting $\frac{Sim\left(i,j\right)}{\sum_{b=1}^{|\mathcal{N}(j)|}S_{bj}}$--the
neighbor weight at step $n$, $\forall i,j:$

\[
\frac{\epsilon}{K\epsilon+2K-2}\leq B_{ij}\left(n\right)\leq\frac{2+\epsilon}{K\epsilon+2}
\]
That yields $\sum_{n=1}^{\infty}\lambda\left(n\right)\geq\frac{\epsilon}{K\epsilon+2K-2}\sum_{n=1}^{\infty}\beta\left(n\right)$
and $\sum_{n=1}^{\infty}\lambda^{2}\left(n\right)\leq\left(\frac{2+\epsilon}{K\epsilon+2}\right)^{2}\sum_{n=1}^{\infty}\beta^{2}\left(n\right)$.
Hence the writing updates converge when $\sum_{n=1}^{\infty}\beta\left(n\right)=\infty$
and $\sum_{n=1}^{\infty}\beta^{2}\left(n\right)<\infty$. We can always
choose such $\beta$ (e.g., $\beta\left(n\right)=\frac{1}{n+1}$). 

With a constant writing rate $\beta$ ($\beta=1/2$), we rewrite Eq.
\ref{eq:write2} as

\begin{align*}
\mathtt{M}_{i}\left(n+1\right) & =\mathtt{M}_{i}\left(n\right)+\beta B_{ij}\left(n\right)\left(\mathtt{G}_{j}\left(n\right)-\mathtt{M}_{i}\left(n\right)\right)\\
 & =\beta B_{ij}\left(n\right)\mathtt{G}_{j}\left(n\right)+\text{\ensuremath{\mathtt{M}_{i}\left(n\right)\left(1-\beta B_{ij}\left(n\right)\right)}}\\
 & =\sum_{t=1}^{n}\beta B_{ij}\left(t\right)\prod_{l=t+1}^{n}\left(1-\beta B_{ij}\left(l\right)\right)\mathtt{G}_{j}\left(t\right)\\
 & +\prod_{t=1}^{n}\left(1-\beta B_{ij}\left(t\right)\right)\mathtt{M}_{i}\left(1\right)
\end{align*}
where the second term $\prod_{t=1}^{n}\left(1-\beta B_{ij}\left(t\right)\right)\mathtt{M}_{i}\left(1\right)\rightarrow0$
as $n\rightarrow\infty$ since $B_{ij}\left(t\right)$ and $\beta$
are bounded between 0 and 1. The first term can be decomposed into
three terms

\begin{align*}
\sum_{t=1}^{n}\beta B_{ij}\left(t\right)\prod_{l=t+1}^{n}\left(1-\beta B_{ij}\left(l\right)\right)\mathtt{G}_{j}\left(t\right) & =T_{1}+T_{2}+T_{3}
\end{align*}
where 
\begin{align*}
T_{1} & =\sum_{t=1}^{n}\beta B_{ij}\left(t\right)\prod_{l=t+1}^{n}\left(1-\beta B_{ij}\left(l\right)\right)Q_{i}\\
T_{2} & =\sum_{t=1}^{n}\beta B_{ij}\left(t\right)\prod_{l=t+1}^{n}\left(1-\beta B_{ij}\left(l\right)\right)\Delta Q_{ij}\left(t\right)\\
T_{3} & =\sum_{t=1}^{n}\beta B_{ij}\left(t\right)\prod_{l=t+1}^{n}\left(1-\beta B_{ij}\left(l\right)\right)\mathtt{\hat{G}}_{j}\left(t\right)
\end{align*}
Here,$Q_{i}$ is the true action value of the state-action stored
in slot $i$, $\Delta Q_{ij}\left(t\right)=Q_{j}\left(t\right)-Q_{i}$
and $\mathtt{\hat{G}}_{j}\left(t\right)=\mathtt{G}_{j}\left(t\right)-Q_{j}\left(t\right)$
the noise term between the return and the true action value. Assume
that the action value is associated with zero mean noise and the action
value noise is independent with the neighbor weights, then $\mathbb{E}\left(T_{3}\right)=0$. 

Further, we make other two assumptions: (1) the neighbor weights are
independent across update steps; (2) the probability $p_{j}$ of visiting
a neighbor $j$ follows the same distribution across update steps
and thus, $\mathbb{E}\left(B_{ij}\left(t\right)\right)=\mathbb{E}\left(B_{ij}\left(l\right)\right)=\mathbb{E}\left(B_{ij}\right)$.
We now can compute 

\begin{align*}
\mathbb{E}\left(T_{1}\right) & =\mathbb{E}\left(\sum_{t=1}^{n}\beta B_{ij}\left(t\right)\prod_{l=t+1}^{n}\left(1-\beta B_{ij}\left(l\right)\right)Q_{i}\right)\\
 & =Q_{i}\sum_{t=1}^{n}\beta\mathbb{E}\left(B_{ij}\right)\prod_{l=t+1}^{n}\left(1-\alpha\mathbb{E}\left(B_{ij}\right)\right)\\
 & =Q_{i}\beta\mathbb{E}\left(B_{ij}\right)\sum_{t=1}^{n}\left(1-\beta\mathbb{E}\left(B_{ij}\right)\right)^{n-t}\\
 & =Q_{i}\beta\mathbb{E}\left(B_{ij}\right)\frac{1-\left(1-\beta\mathbb{E}\left(B_{ij}\right)\right)^{n}}{1-\left(1-\beta\mathbb{E}\left(B_{ij}\right)\right)}\\
 & =Q_{i}\left(1-\left(1-\beta\mathbb{E}\left(B_{ij}\right)\right)^{n}\right)
\end{align*}
As $n\rightarrow\infty$, $\mathbb{E}\left(T_{1}\right)\rightarrow Q_{i}$
since since $B_{ij}\left(t\right)$ and $\alpha$ are bounded between
0 and 1. 

Similarly, $\mathbb{E}\left(T_{3}\right)=\mathbb{E}\left(Q_{j}\left(t\right)-Q_{i}\right)=\mathbb{E}\left(Q_{j}\left(t\right)\right)-Q_{i}=\sum_{j=1}^{|\mathcal{N}(i)|}p_{j}Q_{j}-Q_{i}$,
which is the approximation error of the KNN algorithm. Hence, with
constant learning rate, on average, the $\mathtt{update}$ operator
leads to the true action value (the expected return) plus the approximation
error of KNN. The quality of KNN approximation determines the mean
convergence of $\mathtt{update}$ operator. Since the bias-variance
trade-off of KNN is specified by the number of neighbors $K$, choosing
the right $K>1$ (not too big, not too small) is important to achieve
good writing to ensure fast convergence. That explains why our writing
to multiple slots ($K>1$) is generally better than the traditional
writing to single slot ($K=1$).

\end{document}